%% file: main.tex
\documentclass[10pt,twocolumn,letterpaper]{article}

\usepackage{wacv}              

\input{preamble}

\usepackage{float}

\definecolor{wacvblue}{rgb}{0.21,0.49,0.74}
\usepackage[pagebackref,breaklinks,colorlinks,allcolors=wacvblue]{hyperref}

\def\wacvPaperID{1651} 
\def\confName{WACV}
\def\confYear{2026}
\newcommand{\DeCon}{\textbf{DeCon}}
\newcommand{\DeConSL}{\textbf{DeCon-SL}}
\newcommand{\DeConML}{\textbf{DeCon-ML}}

\title{Beyond the Encoder: Joint Encoder-Decoder Contrastive Pre-Training Improves Dense Prediction}
\usepackage{lipsum}

\makeatletter
\newcommand\customfootnotetext[2]{%
  \begingroup
  \renewcommand\thefootnote{#1}%
  \footnotetext{#2}%
  \endgroup
}
\makeatother

\author{
S\'ebastien Quetin$^{*1}$,
Tapotosh Ghosh$^{*2}$,
Farhad Maleki$^{2}$\\
$^{1}$McGill University, $^{2}$University of Calgary, Canada \\
}
\begin{document}
\maketitle
\customfootnotetext{*}{Equal contributions. Co-first authors are ordered in alphabetical order. Corresponding author: T. Ghosh (tapotosh.ghosh@ucalgary.ca)}

\input{our_sections/0_abstract} 
\input{our_sections/1_intro}

\input{our_sections/2_review} 
\input{our_sections/3_method} 
\input{our_sections/4_results_wacv} 
\input{our_sections/5_discussion_wacv}
\input{our_sections/6_conclusion}
\input{our_sections/acknowledgments} 
{
    \small
    \bibliographystyle{ieeenat_fullname}
    \bibliography{main}
}

\end{document}


\pagestyle{empty}
\thispagestyle{empty}
\maketitle
\pagestyle{empty}
\thispagestyle{empty}


\input{supplementary}

{
    \small
    \bibliographystyle{ieeenat_fullname}
    \bibliography{main}
}

%% file: preamble.tex
\usepackage{pifont}  
\usepackage{xcolor}  
\usepackage{booktabs}
\usepackage{multicol}
\usepackage{multirow}
\usepackage{rotating}
\definecolor{darkgreen}{rgb}{0.0, 0.5, 0.0}
\usepackage{graphicx} 
\newcommand{\cmark}{\ding{51}} 
\newcommand{\xmark}{\ding{55}} 



%% file: our_sections/0_abstract.tex
\begin{abstract}

Contrastive learning methods in self-supervised settings have primarily focused on pre-training encoders, while decoders are typically introduced and trained separately for downstream dense prediction tasks. However, this conventional approach overlooks the potential benefits of jointly pre-training both encoder and decoder. In this paper, we propose \textbf{DeCon}, an efficient encoder-decoder self-supervised learning (SSL) framework that supports joint contrastive pre-training. We first extend existing SSL architectures to accommodate diverse decoders and their corresponding contrastive losses. Then, we introduce a weighted encoder-decoder contrastive loss with non-competing objectives to enable the joint pre-training of encoder-decoder architectures. By adapting a contrastive SSL framework for dense prediction, \textbf{DeCon} establishes consistent state-of-the-art performance on most of the evaluated tasks when pre-trained on Imagenet-1K, COCO and COCO+. Notably, when pre-training a ResNet-50 encoder on COCO dataset, \textbf{DeCon} improves COCO object detection and instance segmentation compared to the baseline framework by +0.37 AP and +0.32 AP, respectively, and boosts semantic segmentation by +1.42 mIoU on Pascal VOC and by +0.50 mIoU on Cityscapes. These improvements generalize across recent backbones, decoders, datasets, and dense tasks beyond segmentation and object detection, and persist in out-of-domain scenarios, including limited-data settings, demonstrating that joint pre-training significantly enhances representation quality for dense prediction. Code is available at \url{https://github.com/sebquetin/DeCon.git}.


\end{abstract}

%% file: our_sections/1_intro.tex
\section{Introduction}
\label{sec:intro}
The growing demand for efficient Deep Learning (DL) methods stems from their ability to address complex tasks. While traditional approaches like supervised learning can train high-performing models, they depend on large volumes of high-quality annotated data. Obtaining such annotations is often labor-intensive, expensive, and sometimes impractical, even when abundant data is available. 
A common workaround is to use pre-trained models, typically encoders trained from scratch on large-scale datasets, to boost performance on downstream tasks with limited labels. Typically, an encoder is pre-trained to capture a representation of the input data, which can then be fine-tuned for downstream tasks such as classification or segmentation. During the downstream fine-tuning phase, randomly initialized layers are typically added on top of the encoder, and the complete architecture is retrained using a supervised learning approach on an annotated dataset. ImageNet~\cite{imagenet} is a widely utilized labeled dataset for such pre-training. However, models pre-trained on ImageNet classification task often transfer poorly to downstream dense prediction tasks, such as detection and segmentation~\cite{imgenet_transfer_to_coco_paper, densecl_paper_ssl, transfusion_paper}. 

Creating large-scale annotated datasets for every task is impractical. Self-Supervised Learning (SSL) is a promising alternative, enabling models to learn from the large-scale unannotated data to enhance performance on various downstream tasks. These pre-trained models can reach downstream performance close to a fully supervised approach while using significantly fewer labeled examples. Notable approaches such as SimCLR~\cite{chen2020simple}, VicReg~\cite{vicreg_paper_ssl}, and MoCo~\cite{moco_paper_ssl} showed promising results by leveraging contrastive learning for downstream classification. However, most SSL frameworks primarily target classification-based downstream tasks. Some efforts, such as PixCon~\cite{pixcon_paper_ssl} and DenseCL~\cite{densecl_paper_ssl}, focus on dense prediction tasks like segmentation or object detection. However, these approaches only pre-train encoders for downstream use and rely on local losses adapted from classification-oriented SSL methods without jointly training the decoder.

In this work, we propose a \textbf{Decoder-aware contrastive learning (DeCon)} approach, an encoder-decoder framework that jointly pre-trains encoder and decoder in an SSL manner. \DeCon\ enhances the representation power of the encoder, and prepares it more effectively for integration with decoders in downstream dense prediction tasks. The contributions of this paper are as follows:

\begin{itemize}
    \item We propose \DeConSL, a single-level joint encoder-decoder adaptation of contrastive dense SSL frameworks. We show that jointly pre-training the encoder and decoder enhances the representation power of the encoder and improves downstream dense prediction performance, even when transferring only the pre-trained encoder.
    
    \item We further extend this adaptation by introducing \DeConML, which builds on \DeConSL\ with a multi-level decoder loss function and an encoder-level channel dropout to promote a comprehensive encoder feature utilization. 
    
    \item \DeCon\ achieves new state-of-the-art (SOTA) results in COCO object detection and instance segmentation, and Pascal VOC, Cityscapes and ADE20K semantic segmentation when pre-trained on ImageNet-1K and COCO+.
    


    \item We show that \DeCon\ consistently improves performance across different backbones, a range of dense downstream tasks beyond object detection, semantic and instance segmentation as well as in various out-of-domain scenarios.

    \item We show that \DeCon\ can achieve gains without increasing parameter count and with comparable GPU cost relative to the original framework, e.g. SlotCon~\cite{slotcon_paper_ssl}.
        
\end{itemize}

%% file: our_sections/2_review.tex
\section{Related Work}
\label{sec:review}

\textbf{Self-Supervised pre-training:} In early SSL works, deep learning models were trained on pretext tasks where supervision was derived directly from the data itself. This enabled the models to learn meaningful representations without relying on manual annotations.
These tasks included predicting missing parts of an image~\cite{impainting}, context prediction~\cite{context_prediction_ssl}, solving jigsaw puzzles~\cite{jigsaw_puzzle}, colorizing grayscale images~\cite{colorization_paper_ssl}, and predicting rotations~\cite{rotation_ssl}. These methods aimed to create surrogate objectives that encouraged models to develop feature representations transferable to downstream tasks.
However, they often produced task-specific representations with limited generalizability.

Recent SSL methods prioritize generalizability and move away from hand-crafted pretext tasks. These approaches can be broadly categorized into generative and contrastive methods. Generative approaches learn representations by modeling the underlying data distribution. This is typically achieved through reconstruction-based methods, where the model predicts corrupted parts of the input~\cite{mae_ssl, jepa_ssl, peco_paper_ssl, simmim_paper_ssl}, or through adversarial methods, where a generator model learns to synthesize realistic samples to challenge a discriminator, thereby encouraging the encoder to extract informative semantic features~\cite{diffusion_ssl_paper, second_diffusion_ssl_paper}. Contrastive methods, which are the focus of this research, learn data representations by maximizing the similarity between features extracted from two augmented views of the same image. To avoid model collapse---a phenomenon where the model maps all inputs to the same representation---strategies such as negative sampling~\cite{moco_paper_ssl,simclr_paper_ssl}, stop-gradient mechanisms~\cite{simsiam_paper}, variance regularization~\cite{vicreg_paper_ssl, vicregl_paper_ssl}, momentum encoders~\cite{moco_paper_ssl}, and other architectural and optimization techniques have been employed~\cite{dino_paper_ssl}.

\noindent\textbf{Dense pre-training:} 
 Although most of the aforementioned frameworks can be used for downstream segmentation or detection, they are not designed specifically for pixel-level---i.e., dense prediction downstream tasks such as instance and semantic segmentation. Consequently, developing SSL approaches that generalize well to such downstream tasks remains an active area of research~\cite{slotcon_paper_ssl,densecl_paper_ssl,cp2_paper_ssl,vicregl_paper_ssl,vader_paper_ssl}. The main idea behind these approaches is to shift from a global image similarity to a local similarity. They aim to enforce local similarity based on the location of pixels in an image~\cite{pixpro_paper_ssl, cp2_paper_ssl,vader_paper_ssl}, local pixel features~\cite{densecl_paper_ssl}, or both~\cite{10030381, vicregl_paper_ssl, pixcon_paper_ssl}. Local similarity can also be enforced in a region-based manner~\cite{ge2023soft, henaff2022object, NEURIPS2021_f1b6f285, slotcon_paper_ssl}. These methods outperform those relying solely on image-level global similarity when evaluated on dense prediction downstream tasks.
 
\noindent\textbf{Encoder-only versus encoder-decoder pre-training:} 
Encoder-decoder architectures with skip connections between the encoder and decoder layers (i.e., the U-Net family of models) have achieved state-of-the-art results in many segmentation tasks~\cite{unet_paper, nnunet_paper,unet++_paper}. A common approach for pre-training such models in a self-supervised manner is image reconstruction~\cite{image_reconstruction_unet1, Brempong_2022_CVPR, CHEN2019101539}. However, the presence of skip connections allows a substantial amount of information to bypass the encoder's final layer---often referred to as the bottleneck layer--resulting in lower-quality embeddings generated by the encoder. Our proposed approach preserves skip connections while addressing the issue of inferior representations generated by the decoder.

Encoder-only contrastive learning frameworks are built solely around an encoder, incorporating components such as predictors and projectors. However, they do not include a decoder. After pre-training, a randomly initialized decoder is attached to the pre-trained encoder in dense prediction downstream tasks to form a complete architecture. 
While some generative SSL frameworks use a decoder for pre-training~\cite{Dai_2021_CVPR, impainting, mae_ssl}, a clear gap remains in exploring vision model pre-training that jointly employs encoder and decoder losses within a unified contrastive learning framework to learn informative data representations.



%% file: our_sections/3_method.tex
\begin{figure*}[!h]
  \centering
   \includegraphics[width=0.90\textwidth]{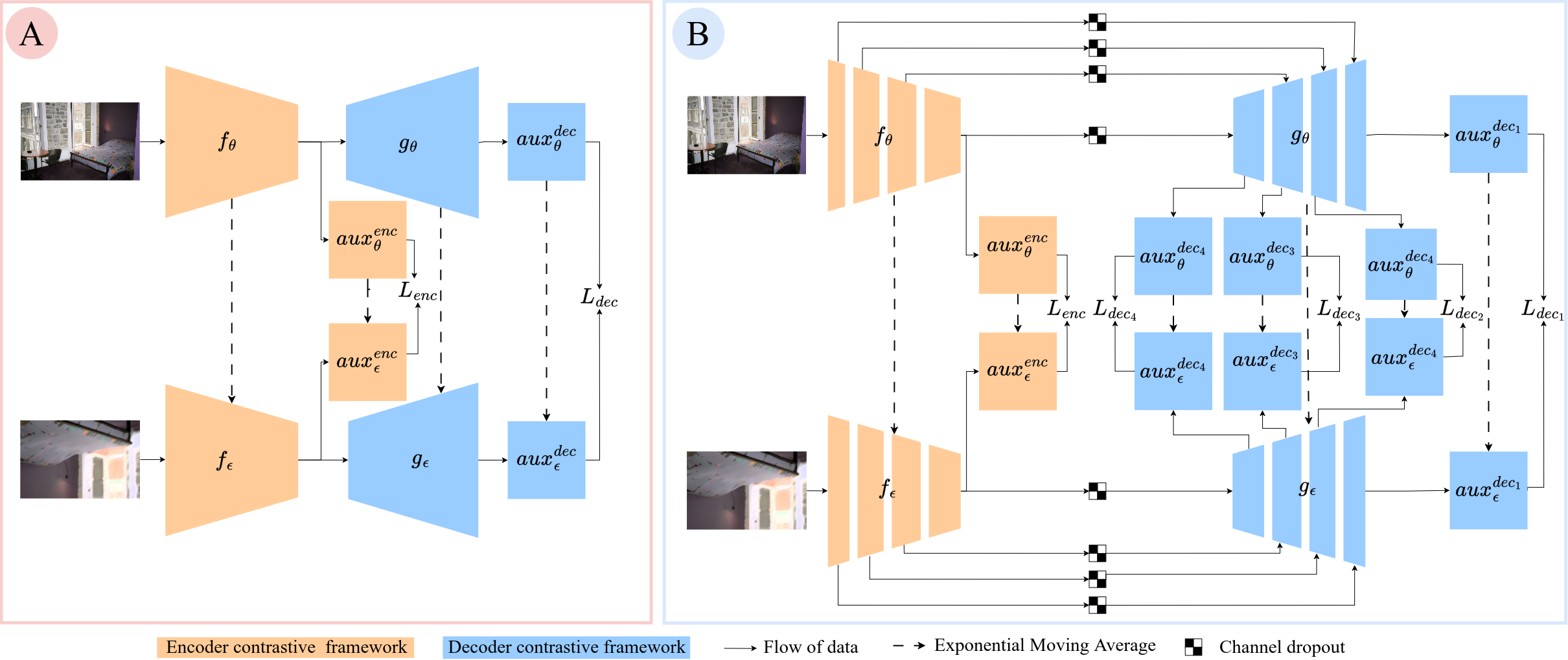}
   \caption{\textbf{A: DeCon-SL}. Instead of a classical encoder-only pre-training, a decoder is pre-trained alongside the encoder. We mirror the encoder loss at the decoder level and optimize the architecture using a weighted sum of encoder and decoder losses. \textbf{B: DeCon-ML}. Instead of computing the decoder loss at a single level, it is calculated across multiple levels (four in this figure). Additionally, a channel-wise dropout is applied at the output of each encoder level before it is passed through the skip connection to the decoder. }
   \label{fig:combined-framework}
\end{figure*}
\section{Methods}
\label{sec:methods}
We introduce two architectural adaptations, as depicted in~\cref{fig:combined-framework}: (A) a single-level decoder loss (\DeConSL) and (B) a multi-level decoder loss (\DeConML).\par

\subsection{\DeConSL}
Given a vision SSL framework that pre-trains an encoder using a teacher-student architecture~\cite{hintonTeacherStudent}, we retain the encoders and any existing ``auxiliary layers'' (e.g., predictors, projection heads), and add decoders and their corresponding auxiliary layers for both teacher and student networks. For all models, the decoder auxiliary layers are designed to match those of the encoder, with adjustments applied to align with the decoder's output feature size. We obtain two losses, one is computed from the encoder features while the other one is computed from the decoder features. Consequently, the loss function is defined as follows:
\begin{equation}
  Loss = \alpha \times L_{enc} + (1 - \alpha) \times L_{dec}
  \label{eq:loss}
\end{equation}


\noindent where, $\alpha$ represents the contribution of the encoder loss in the objective function.\par
The DeCon-SL architecture is illustrated in \cref{fig:combined-framework} A, where $f_\theta$ and $f_\epsilon$ represent student and teacher encoders, respectively. $aux^{enc}_{\theta} $ and $aux^{enc}_{\epsilon} $ are the SSL framework's auxiliary layers  of the student and teacher encoders, respectively. Auxiliary layers can include any projector, predictor heads, or any layers used in the loss definition of the original SSL framework. $g_\theta$ and $g_\epsilon$ are the decoders of the student and teacher networks, and $aux^{dec}_{\theta} $ and $aux^{dec}_{\epsilon} $ their respective auxiliary layers. $g_\theta$ and $g_\epsilon$ take the output of encoder $f_\theta$ and $f_\epsilon$ as input, respectively. Auxiliary layers $aux^{enc}_{\theta} $ and $aux^{enc}_{\epsilon} $ are responsible for generating a representation from the encoder output, that is used to compute a contrastive loss for the encoders $ L^{encoder}$. The contrastive loss, $ L^{decoder}$ is calculated from the representation obtained from the auxiliary layers of the decoders, i.e.,  $aux^{dec}_{\theta} $ and $aux^{dec}_{\epsilon}$. $f_\epsilon$ and $g_\epsilon$ are updated using the exponential moving average (EMA) of $f_\theta$ and $g_\theta$, respectively.


\subsection{DeCon-ML}
To promote a comprehensive usage of the encoder's parameters, we introduce a multi-layer contrastive decoder loss. This updated framework consists of a single loss for the encoder and multiple loss components for the decoder in a deep supervision manner along with channel dropout, which we describe in detail below. \Cref{fig:combined-framework} B illustrates the DeCon-ML adaptation of a given SSL framework.

\noindent\textbf{Channel dropout:}
The dropout layer is typically used to prevent overfitting, improve generalization, and act as an implicit ensemble~\cite{classic_dropout_paper,channel_dropout_paper}. However, we extend its use by applying dropout to channels transferred through skip connections between the encoder and decoder. In DeCon-ML, we apply channel dropout to the output layers of the different encoder levels  by zero-ing out entire channels of the feature maps when passing these outputs to the decoder. No channel is zeroed-out when the inputs pass through the encoder. This channel dropout prevents the model from over-relying on specific features shared through skip connections. It encourages a more comprehensive use of the encoder’s parameters at the different levels, leading to a richer and more powerful data representation learned, while still preserving the input information content as it passes through the encoder to its bottleneck.

\noindent\textbf{Decoder deep supervision:}
An encoder-decoder SSL framework allows us to retrieve meaningful features from multiple levels of the decoder. We can then pre-train the architecture with multiple losses using decoder deep supervision. If the decoder is connected to the encoder at multiple levels through lateral/skip connections, decoder deep supervision strengthens the representation power of the encoder at different levels~\cite{deep_sup_paper}. Since a decoder passes the encoder bottleneck representation in a bottom-up manner to its upper level layers, the final encoder bottleneck representation remains a non-negligible objective of the training since it impacts all decoder level losses. In contrast, applying deep supervision at multiple encoder stages in an encoder-only SSL framework might prevent the creation of a strong semantic representation at the encoder bottleneck. To perform the decoder deep supervision, we add auxiliary layers to each decoder level, and a loss is computed at each level between the teacher and student representations. When performing decoder deep supervision, the encoder loss will remain the same as~\cref{eq:loss}, but the decoder loss will be updated as follows:
%
\begin{align}
Loss = & \alpha \times L_{enc} + (1 - \alpha) \times L_{dds}\label{eq:totalddsloss}\\
  L_{dds} = & \frac{1}{j} \times \sum_{i=1}^{j} L_{dec_i}
  \label{eq:ddsloss}
\end{align}
\noindent where $j$ is the number of decoder levels at which a loss will be calculated and $L_{dec_i}$ is the loss computed at level $i$. The final decoder loss in \cref{eq:totalddsloss} is thus an average of the losses at the different levels.

\subsection{Implementation details}

\textbf{Architecture: }
We use a ResNet-50 encoder~\cite{resnet_paper} as a backbone and implement a Fully Convolutional Network (FCN)~\cite{fcn_paper} and a skip-connected Feature Pyramid Network (FPN)~\cite{fpn_paper} as two decoder architectures in the DeCon framework. 
We follow the \textit{mmseg}~\cite{mmseg_package} FCN implementation from SlotCon~\cite{slotcon_paper_ssl} and customize the \textit{Detectron2}~\cite{detectron2_package} FPN implementation. Details on these decoders are provided in Supplementary Material Section 1. We use DeCon to adapt the SlotCon~\cite{slotcon_paper_ssl} framework. Additional experiments are conducted using DenseCL~\cite{densecl_paper_ssl} and PixPro~\cite{pixpro_paper_ssl} framework, or a ConvNeXt-Small~\cite{convnext_paper} backbone. 
For SlotCon, the student and teacher encoders, and every auxiliary layers---i.e., the respective projectors and semantic groupings, the slot predictor and the prototypes---are kept unchanged. Two decoders, a student and a teacher decoder (either FCN or FPN), that take the output of their corresponding encoder as input are added to the framework. The auxiliary layers are replicated at the decoder level. In the case of decoder deep supervision with FPN decoders (DeCon-ML), auxiliary layers are replicated independently at the four decoder levels. No deep supervision is performed in the DeCon adaptation of SlotCon with an FCN decoder (DeCon-SL). The decoder projectors differ from the encoder projectors only in input channels: 2048 for the encoder and 256 for the FPN and FCN decoders. We follow SlotCon by updating the student branch and the corresponding encoder and decoder slot predictors with backpropagation, the teacher branch with an EMA of the first one, and the prototypes centers with an EMA of the previous batch centers seen during training. Our implementation is based on the official SlotCon repository. 

We adapt in a similar way DenseCL and PixPro into a DeCon-SL framework using an FCN decoder. More details are given in Supplementary Material Section 2. 
We adapt DenseCL implementation from \textit{mmselfsup} package~\cite{mmselfsup2021} version 1.0.0 and PixPro from the original implementation.

Although the pre-trained backbone remains unchanged, adding multiple decoder levels and associated loss functions increases the overall number of parameters in the DeCon-ML pre-training. To enable a fair comparison with the original SlotCon framework using a similar parameter budget, we implemented a reduced version of DeCon-ML that includes only the first two decoder levels originating from the encoder bottleneck. Additionally, the hidden dimension in the decoder projector was reduced from 4096 to 2048. This smaller version, which matches the parameter count of the original SlotCon framework (when $\alpha = 0$), is referred to as \textbf{DeCon-ML-S}, while the full model with four decoder levels and unmodified configuration is denoted \textbf{DeCon-ML-L}. Detailed figures of DeCon-SL, DeCon-ML-S and DeCon-ML-L adaptations of SlotCon are available in the Supplementary Material Section 3.



\noindent\textbf{Pre-training setup:} We pre-train DeCon on COCO 2017 dataset~\cite{coco_dataset}, COCO+ (\textit{train2017} + \textit{unlabeled2017} datasets) and ImageNet-1K datasets~\cite{imagenet}. The augmentations and hyperparameters---learning rate (LR), weight decay, schedulers, input shape, momentum, and so on---are kept the same as the adapted SSL framework.
%
Models are pre-trained on COCO and COCO+ for 800 epochs, and for 200 epochs on ImageNet-1K.
The SlotCon ResNet-50 models are pre-trained using a LARS optimizer, with a total batch size of 512 split across all available GPUs. The base LR is 1.0 scaled with the batch size following~\cite{slotcon_paper_ssl} and updated with a cosine LR decay schedule using a weight decay of $10^{-5}$ and 5 warm-up epochs. The ConvNeXt-Small backbone is pre-trained on COCO dataset with SlotCon and DeCon-ML using an AdamW optimizer~\cite{adamw_opt_paper}, a base LR of 0.002 and weight-decay 0.05. To compare with ViT-based methods, we also pre-train a ConvNeXt-S backbone on ImageNet-1K for 250 epochs with DeCon-SL framework and an FCN decoder using a base LR of 0.001 and drop path rate of 0.1. Remaining hyperparameters are kept the same as for the ResNet-50 implementation. The pre-trainings using ConvNeXt-S backbones or FPN decoders are done using eight 80GB NVIDIA H100 GPUs. The pre-trainings without decoder and with FCN decoders are conducted on one 80GB NVIDIA A100 GPU.
%
%

To evaluate the generalization of our approach to different SSL frameworks, we adapt DenseCL and PixPro frameworks with DeCon-SL by pre-training a ResNet-50 encoder with an FCN decoder on COCO 2017 dataset. Optimization follows the base frameworks. To evaluate the dependence to the pre-training dataset, we also pre-train a similar DeCon-SL adaptation of SlotCon on REFUGE~\cite{dataset_refuge} and ISIC~\cite{dataset_isic} datasets. We provide all pre-training details in Supplementary Material Section 2.

\subsection{Evaluation protocol}
\label{sec:eval_protocol}
We evaluate the different pre-trained models on various downstream tasks. Unless stated otherwise, we only transfer the pre-trained encoder for the fine-tuning experiments. We initialize the network with a pre-trained encoder and fine-tune end-to-end in all cases. All fine-tuning results reported are averaged over three independent runs. The pre-trainings, however, are only run once. All fine-tuning experiments related to the same downstream task are run on the same hardware. We perform semantic segmentation tasks using \textit{mmsegmentation}~\cite{mmseg_package} version 0.30.0. and other remaining tasks using \textit{Detectron2} package~\cite{detectron2_package} version 0.6.

\noindent\textbf{Object detection and instance segmentation:} We fine-tune a Mask R-CNN with a pre-trained ResNet-50 or ConvNeXt-Small backbone on COCO 2017 dataset~\cite{coco_dataset}.

\noindent\textbf{Semantic segmentation:}
We fine-tune a pre-trained ResNet-50 or ConvNeXt-Small with an FCN decoder on Pascal VOC~\cite{voc_dataset}, and ResNet-50 with an FCN decoder on ADE20K~\cite{ade} and Cityscapes~\cite{cityscape_dataset} following SlotCon~\cite{slotcon_paper_ssl} and PixCon~\cite{pixcon_paper_ssl}. We also fine-tune an ImageNet-1K-pre-trained ConvNeXt-Small backbone with an UPerNet decoder on ADE20K dataset. 

\noindent\textbf{Panoptic segmentation, keypoint detection, and dense pose estimation:} We adapt task-specific \textit{Detectron2} configurations based on the SlotCon object detection architecture, and fine-tune on the COCO 2017 dataset for panoptic segmentation and keypoint detection, and on the COCO 2014 dataset for dense pose estimation.

\noindent\textbf{Generalization to out-of-domain datasets:} We fine-tune a ResNet-50-FCN architecture on REFUGE~\cite{dataset_refuge} and ISIC 2017~\cite{dataset_isic} datasets for medical semantic segmentation using 5, 25, and 100\% of the training data. We also fine-tune a ResNet-50-deeplabv3+ architecture for PlantSeg ~\cite{plantseg} semantic segmentation and a Faster R-CNN ResNet-50-FPN architecture for object detection on PlantDoc~\cite{plantdoc} and Detecting Disease ~\cite{detecting-diseases_dataset} datasets. Fine-tuning on agriculture datasets is performed using 10 and 100\% of the datasets. 


More details on the training hyperparameters and target datasets are provided in Supplementary Material Section 2.

%% file: our_sections/4_results_wacv.tex
\section{Results}
\label{sec:results}

\subsection{Experiments}

\begin{table*}[htbp]
  \centering
  \caption{Performance of DeCon-SL ($\alpha=0.25$) and DeCon-ML (Small/Large, $\alpha=0$, dropout=0.5) adaptations of SlotCon with a ResNet-50 backbone pre-trained for 800 epochs on COCO/COCO+ and 200 epochs on ImageNet-1K, then fine-tuned on different downstream tasks. Only pre-trained encoders were transferred. PixCon ImageNet-1K results are not reported due to unavailable checkpoints. Results are averaged over three runs. \textsuperscript{\textdagger} Re-implemented, * Results from original paper, $\lozenge$ Result from SlotCon \cite{slotcon_paper_ssl} paper.}
 \scalebox{0.60}{
\begin{tabular}{clccccccccccc||cccc}
\toprule
\multicolumn{1}{c}{\multirow{3}[4]{*}{\textbf{Pret. Dataset}}} &
\multicolumn{1}{c}{\multirow{3}[4]{*}{\textbf{Framework}}}  &
\multicolumn{1}{c}{\multirow{3}[4]{*}{\textbf{Backbone}}} &
\multicolumn{1}{c}{\multirow{3}[4]{*}{\textbf{Pret. Dec.}}} &
\multicolumn{3}{c}{\textbf{Object Det.}} &
\multicolumn{3}{c}{\textbf{Instance Seg.}} &
\multicolumn{3}{c}{\textbf{Semantic Seg.}} & \multicolumn{2}{c}{\textbf{Dense pose Est.}} & \textbf{Panoptic Seg.} & \textbf{Keypoint Det.} \\
& & & &
\multicolumn{3}{c}{\textbf{COCO}} &
\multicolumn{3}{c}{\textbf{COCO}} &
\textbf{VOC} & \textbf{City} & \textbf{ADE} & \multicolumn{2}{c}{\textbf{COCO}} & \textbf{COCO} & \textbf{COCO} \\
\cmidrule(lr){5-7} \cmidrule(lr){8-10} \cmidrule(lr){11-11} \cmidrule(lr){12-12} \cmidrule(lr){13-13} \cmidrule(lr){14-15} \cmidrule(lr){16-16} \cmidrule(lr){17-17}
& & & &
\textbf{AP} & \textbf{AP50} & \textbf{AP75} &
\textbf{AP} & \textbf{AP50} & \textbf{AP75} &
\textbf{mIoU} & \textbf{mIoU} & \textbf{mIoU} & \textbf{GPS AP} & \textbf{GPSm AP} & \textbf{PQ} & \textbf{AP} \\
\midrule
    
\textbf{-} & Random init.$\lozenge$ & ResNet-50 & \textbf{-} & 32.8  & 50.9  & 35.3  & 29.9  & 47.9  & 32.0  & 39.5  & 65.3 & 29.4 & 60.68 & 64.45 & 33.51 & 63.92 \\
\midrule
\multirow{7}[1]{*}{COCO}
& PixCon (2024)~\cite{pixcon_paper_ssl}* & \multirow{5}[1]{*}{ResNet-50} & \textbf{-}     & 40.81 & 60.97  & 44.80  & 36.80 & 57.93  & 39.62  & \underline{72.95} & \textbf{76.62} & 38.00 &\textbf{-} &\textbf{-} &\textbf{-} &\textbf{-} \\
  & SoCo-D (2024) \cite{mindAug2024}* && \textbf{-} & 40.3  & 60.1  & 44.0  & 35.1  & 56.9    & 37.6  & \textbf{-}  & \textbf{-} & \textbf{-} &\textbf{-} &\textbf{-}  &\textbf{-} &\textbf{-} \\
& SlotCon (2022)~\cite{slotcon_paper_ssl}\textsuperscript{\textdagger} && \textbf{-}     & 40.81 & 60.95 & 44.37 & 36.80 & 57.98 & 39.54 & 71.50 & 75.95 & 38.57 & 63.39 & 64.84 & 40.31 & 65.66\\
& DeCon-SL && FCN   & 40.97 & 61.22 & 44.81 & 36.92 & 58.12 & 39.78 & \underline{73.01} & 76.21 & \textbf{38.81} &63.50&64.98&40.44&65.77\\
& DeCon-ML-S && FPN   & 40.97 & 61.20 & 44.71 & 36.94 & 58.20 & 39.63 & 72.80 & 76.21 & 38.36 &\textbf{63.75}&\textbf{65.08}&40.52&65.66\\
& DeCon-ML-L && FPN   & \textbf{41.18} & \textbf{61.38} & \textbf{44.91} & \textbf{37.12} & \textbf{58.35} & \textbf{39.94} & \underline{72.92} & 76.45 & 38.70 & 63.72 & 64.94 & \textbf{40.90} & \textbf{65.88}\\


\midrule
\multirow{4}[1]{*}{COCO+}
& PixCon (2024)~\cite{pixcon_paper_ssl}* & \multirow{4}[1]{*}{ResNet-50} & \textbf{-}     & 41.2  & \textbf{-}     & \textbf{-}     & 37.1  & \textbf{-}     & \textbf{-}     & 73.9  & \multicolumn{1}{c}{\underline{77.00}} & 38.80 &\textbf{-} &\textbf{-} &\textbf{-} &\textbf{-} \\

& SlotCon (2022)~\cite{slotcon_paper_ssl}\textsuperscript{\textdagger} && \textbf{-}     & 41.63 & 62.10 & 45.67 & 37.57 & 59.07 & 40.45 & 73.93 & 76.43 & 39.11&\textbf{-} &\textbf{-}  &\textbf{-} &\textbf{-} \\
& DeCon-SL && FCN   & 41.86 & \underline{62.43} & 45.73 & 37.75 & 59.40 & 40.48 & 74.46 & 76.65 & \textbf{39.25} &\textbf{-} &\textbf{-} &\textbf{-} &\textbf{-} \\
& DeCon-ML-L && FPN   & \textbf{42.08} & \underline{62.42} & \textbf{46.13} & \textbf{37.84} & \textbf{59.41} & \textbf{40.75} & \textbf{75.36} & \underline{77.00} & 39.04 &\textbf{-} &\textbf{-} &\textbf{-} & \textbf{-} \\

\midrule
\multicolumn{1}{c}{\multirow{4}[1]{*}{ImageNet-1K}}  &  Supervised $\lozenge$ & \multicolumn{1}{c}{\multirow{4}[1]{*}{ResNet-50}} & \textbf{-} & 39.7  & 59.5  & 43.3  & 35.9  & 56.6  & 38.6  & 74.4  & 74.6 & 37.9 & \textbf{-} & \textbf{-} & \textbf{-} & \textbf{-} \\
& DINO (2021)~\cite{dino_paper_ssl}\textsuperscript{\textdagger} && \textbf{-}     & 40.24 & 60.25  & 44.13 & 36.47 & 57.49 & 39.20 & 73.09 & 75.57 & 37.30&\textbf{-} & \textbf{-} &\textbf{-} &\textbf{-} \\
& SlotCon (2022)~\cite{slotcon_paper_ssl}\textsuperscript{\textdagger} && \textbf{-}     & 41.69 & 62.07 & 45.59 & 37.59 & 58.97 & 40.49 & 75.02 & 76.15 & 38.97 &\textbf{-} &\textbf{-} &\textbf{-} &\textbf{-} \\
& DeCon-ML-L && FPN   & \textbf{41.80} & \textbf{62.12} & \textbf{45.73} & \textbf{37.73} & \textbf{59.08} & \textbf{40.68} & \textbf{75.40} & \textbf{76.51} & \textbf{39.01}&\textbf{-} &\textbf{-} &\textbf{-} &\textbf{-} \\

\bottomrule
\end{tabular}
}
  \label{tab:main}%
\end{table*}%

\begin{table}[htbp]
  \centering
  \caption{Performance of DeCon-ML-L ($\alpha=0$, dropout = 0.5) and SlotCon with ConvNeXt-S encoder, pre-trained for 800 epochs on COCO and fine-tuned on segmentation and detection tasks.}
    \scalebox{0.6}{
    \begin{tabular}{clcccc}
    \toprule
    \multicolumn{1}{c}{\multirow{3}[4]{*}{\textbf{Pret. Dataset}}} & \multicolumn{1}{c}{\multirow{3}[4]{*}{\textbf{Framework}}} & \multicolumn{1}{c}{\multirow{3}[4]{*}{\textbf{Pret. Dec.}}} & \textbf{Obj. Det.} & \textbf{Inst. Seg.} & \textbf{Sem. Seg.} \\
          &       &       & \textbf{COCO} & \textbf{COCO} & \textbf{VOC} \\
\cmidrule{4-6}          &       &       & \textbf{AP} & \textbf{AP} & \textbf{mIoU} \\
    \midrule
    \multirow{2}[2]{*}{COCO} & SlotCon & \multicolumn{1}{c}{\textbf{-}} & 44.07 & 39.67 & 73.24 \\
          & DeCon-ML-L & FPN   & \textbf{44.71} & \textbf{40.37} & \textbf{73.81} \\
    \bottomrule
    \end{tabular}%
    }
  \label{tab:convnext}%
\end{table}%

\begin{table}[htbp]
  \centering

  \caption{Comparison of ViT and ConvNeXt models pre-trained on ImageNet-1K.  We pre-trained a ConvNeXt-S encoder using DeCon-SL (FCN decoder, $\alpha = 0.25$). All models are fine-tuned with an UPerNet decoder~\cite{upernet} on ADE20K. *Results from MixedAE~\cite{mixedAE}; \textdagger Results using two decoder levels to match with DeCon-SL decoder.}
  
  \scalebox{0.58}{
    \begin{tabular}{lcccc}
    \toprule
    \textbf{Framework} & \textbf{Pret. Backb.} & \multicolumn{1}{c}{\textbf{Backb. Params. (M)}} & \multicolumn{1}{c}{\textbf{Pret. Epoch}} & \multicolumn{1}{c}{\textbf{ADE20k (mIoU)}} \\
    \midrule
    MoCov3 (2021)~\cite{mocov3}* & ViT-B & 88    & 600   & 46.8 \\
    DINO (2021)~\cite{dino_paper_ssl}* & ViT-B & 88    & 1600  & 46.9 \\
    BEiT (2021)~\cite{beit}*  & ViT-B & 88    & 300   & 44.7 \\
    MAE (2022)~\cite{mae_ssl}* & ViT-B & 88    & 300   & 46.7 \\
    MixedAE (2023)~\cite{mixedAE}*\textdagger & ViT-B & 88    & 300   & 47.4 \\
    DeCon-SL [ours] & ConvNeXt-S & 50    & 250   & \textbf{48.02} \\
    \bottomrule
    \end{tabular}
    }%
  \label{tab:generative}%
\end{table}%

\noindent\textbf{ResNet-50 backbone: }\Cref{tab:main} and Supplementary Material Table S1 present the results of fine-tuning a ResNet-50 backbone pre-trained with various contrastive SSL frameworks. We mainly compare with SlotCon and PixCon as they demonstrated the most competitive performance in dense downstream tasks among previously published contrastive methods. DeCon-ML-L is consistently establishing a new SOTA for all tasks when pre-trained on ImageNet-1K and is consistently improving upon its base framework SlotCon and showing SOTA results for most of the evaluated downstream tasks when pre-trained on COCO and COCO+ dataset. Notably, it reduces the gap between PixCon and SlotCon for Pascal VOC and Cityscapes semantic segmentation when pre-trained on COCO. This supports the fact that the DeCon adaptation is robust across pre-training datasets and fine-tuning tasks. Moreover, DeCon-ML-S and DeCon-SL, whose training costs are comparable to SlotCon exhibit better performance than SlotCon and other contrastive frameworks. Finally, \cref{tab:main} also shows that performance improvement compared to SlotCon are robust to other downstream tasks such as dense pose estimation, panoptic segmentation and keypoint detection.


\noindent\textbf{ConvNeXt backbone and comparisons with ViT: }\Cref{tab:convnext} shows the fine-tuning performance of a ConvNeXt-Small backbone pre-trained using DeCon-ML-L and the base SlotCon framework. A consistent improvement is seen with our encoder-decoder framework, confirming the robustness of our method to more modern and bigger backbones. Notably, both the absolute performance and the relative improvement are greater than those observed with the ResNet-50 backbone. \Cref{tab:generative} presents a performance comparison between a ConvNeXt-S model pre-trained on ImageNet-1K with DeCon-SL and ViT backbones pre-trained using various frameworks. Despite its smaller size and shorter pre-training schedule, the ConvNeXt-S backbone, pre-trained for dense tasks, outperforms all competing methods.

\begin{table}[htbp]
  \centering
  \scriptsize
  \caption{Performance of DeCon-SL ($\alpha=0.5$) adapting DenseCL and PixPro with a ResNet-50 encoder, pre-trained for 800 epochs on COCO and fine-tuned on Pascal VOC and Cityscapes.}
  \scalebox{0.75}{
     \begin{tabular}{ccclcc}
    \toprule
    \multicolumn{1}{c}{\multirow{3}[4]{*}{\textbf{Pret. Dataset}}} & \multicolumn{1}{c}{\multirow{3}[4]{*}{\textbf{Framework}}} & \multicolumn{1}{c}{\multirow{3}[4]{*}{\textbf{Pret. Dec.}}} & \multicolumn{1}{c}{\multirow{3}[4]{*}{\textbf{Transfer}}} & \multicolumn{2}{c}{\textbf{Sem. Seg.}} \\
          &       &       &       & \textbf{VOC} & \textbf{City} \\
\cmidrule{5-6}          &       &       &       & \textbf{mIoU} & \textbf{mIoU} \\
    \midrule
    \multirow{6}{*}{COCO} & DenseCL & \multicolumn{1}{c}{-} & \multicolumn{1}{c}{Enc} & 71.11 & 75.87 \\
\cmidrule{2-6}          & \multicolumn{1}{c}{\multirow{2}[2]{*}{DeCon-SL  (DenseCL)}} & \multirow{2}[2]{*}{FCN} & \multicolumn{1}{c}{Enc} & 71.31 & 76.04 \\
          &       &       & Enc + Dec & \textbf{71.59} & \textbf{76.17} \\
\cmidrule{2-6}          & PixPro & \multicolumn{1}{c}{-} & \multicolumn{1}{c}{Enc} & 72.13 & 75.84 \\
\cmidrule{2-6}          & \multicolumn{1}{c}{\multirow{2}[2]{*}{DeCon-SL-(PixPro)}} & \multirow{2}[2]{*}{FCN} & \multicolumn{1}{c}{Enc} & 72.18 & 76.16 \\
          &       &       & Enc + Dec & \textbf{73.08} & \textbf{76.19} \\
    \bottomrule
    \end{tabular}%
    }
  \label{tab:densecl}%
\end{table}%

\noindent\textbf{Generalizability across frameworks: }\Cref{tab:densecl} presents semantic segmentation fine-tuning results for models pre-trained with DenseCL and PixPro, along with their DeCon-SL adaptations. Pre-training with a decoder improves performance over the base frameworks, and transferring the decoder further enhances results, supporting the hypothesis that DeCon can be adapted to multiple frameworks. However, transferring the pre-trained decoder does not consistently benefit downstream performance in the case of SlotCon, as seen in~\cref{tab:ood} and Supplementary Material 5.



\noindent\textbf{Out-of-domain tasks: }
\Cref{tab:ood} shows fine-tuning results of COCO-pre-trained encoders on REFUGE and ISIC. DeCon-SL pre-training consistently outperforms SlotCon, with larger gains under limited data. Transferring both encoder and decoder further improves performance on ISIC. Additionally, pre-training only on ISIC dataset consistently resulted in better performance compared to pre-training on COCO dataset. However, pre-training on REFUGE dataset did not outperform COCO pre-trained approaches, potentially due to the lack of available images (only 400) to develop a strong enough representation during pre-training. In both ISIC and REFUGE pre-trainings, DeCon-SL outperformed SlotCon. \Cref{tab:agri} shows that DeCon variants also perform better than SlotCon in agricultural object detection and semantic segmentation tasks using 10 or 100\% of the  datasets. 
DeCon pre-trained models also outperformed randomly initialized ones.
\begin{table}[htbp]
  \centering
  \caption{Transfer performance with SlotCon and DeCon-SL (ResNet-50, FCN, $\alpha{=}0.5$) pre-trained on COCO, REFUGE, and ISIC, then fine-tuned on out-of-domain medical datasets (REFUGE and ISIC sem. seg. ) in various data settings.}
  \scalebox{0.490}{ 
    \begin{tabular}{ccccc|lll|lll|ll}
    \toprule
    \multicolumn{1}{c}{\multirow{3}[6]{*}{\textbf{\shortstack{SSL\\strategy}}}} & \multicolumn{2}{c}{\multirow{2}[4]{*}{\textbf{Loss}}} & \multicolumn{2}{c}{\multirow{2}[4]{*}{\textbf{Transfer}}} & \multicolumn{6}{c}{\textbf{\shortstack{Pre-trained on\\ COCO}}} & \multicolumn{2}{c}{\textbf{\shortstack{Pre-trained on \\Downstream Data} }} \\
\cmidrule{6-13}          & \multicolumn{2}{c}{} & \multicolumn{2}{c}{} & \multicolumn{3}{c}{\textbf{REFUGE}} & \multicolumn{3}{c}{\textbf{ISIC}} & \textbf{REFUGE} & \multicolumn{1}{p{3em}}{\textbf{ISIC}} \\
          & &  &  &  & \textbf{mIoU} & \textbf{mIoU} & \textbf{mIoU} & \textbf{mIoU} & \textbf{mIoU} & \textbf{mIoU} & \textbf{mIoU} & \multicolumn{1}{p{3em}}{\textbf{mIoU}} \\

\cmidrule{2-13}          & \boldmath{}\textbf{$L_{enc}$}\unboldmath{} & \boldmath{}\textbf{$L_{dec}$}\unboldmath{} & \textbf{enc} & \textbf{dec} & \textbf{5\%} & \textbf{25\%} & \textbf{100\%} & \textbf{5\%} & \textbf{25\%} & \textbf{100\%} & \textbf{100\%} & \multicolumn{1}{p{3em}}{\textbf{100\%}} \\
    \midrule
    \multicolumn{1}{c}{\textbf{Random init.}} & \textbf{-} & \textbf{-} & \textbf{-} & \textbf{-} & 49.53 & 41.41 & 62.82 & 75.66 & 78.38 & 80.65 & 62.82 & 80.65 \\
    \midrule
    \textbf{SlotCon} & \cmark & \xmark & \cmark & \xmark & 69.60 & 77.80 & 82.83 & 74.95 & 79.05 & 82.52 & 79.86 & 83.19 \\
    \midrule
    \multirow{2}[2]{*}{\textbf{DeCon-SL}} & \cmark & \cmark & \cmark & \xmark & 71.75 & \textbf{78.92} & \textbf{83.57} & 75.62 & 79.56 & 82.84 & 80.71 & \textbf{83.66} \\
          & \cmark & \cmark & \cmark & \cmark & \textbf{72.09} & 77.27 & 83.25 & \textbf{76.00} & \textbf{79.97} & \textbf{82.94} & \textbf{81.01} & 83.25 \\
    \bottomrule
    \end{tabular}%
    }
  \label{tab:ood}%
\end{table}

\begin{table}[htbp]
  \centering
  \caption{ResNet-50 backbone transfer performance of SlotCon, DeCon-SL (FCN decoder) and DeCon-ML (FPN decoder),  pre-trained on COCO and fine-tuned on agriculture datasets with 10\% and 100\% labeled data.}
  \scalebox{0.5}{
    \begin{tabular}{lcccccc}
    \toprule
    \multicolumn{1}{c}{\multirow{3}[6]{*}{\textbf{Framework}}} & \multicolumn{2}{c}{\textbf{PlantDoc (Obj. Det.) }} & \multicolumn{2}{c}{\textbf{Detecting Diseases (Obj. Det.)}} & \multicolumn{2}{c}{\textbf{PlantSeg (Sem. Seg.)}} \\
\cmidrule(lr){2-3} \cmidrule(lr){4-5} \cmidrule(lr){6-7}          & \textbf{AP} & \textbf{AP} & \textbf{AP} & \textbf{AP} & \textbf{mIoU} & \textbf{mIoU} \\
\cmidrule(lr){2-3} \cmidrule(lr){4-5} \cmidrule(lr){6-7}          & \textbf{10\%} & \textbf{100\%} & \textbf{10\%} & \textbf{100\%} & \textbf{10\%} & \textbf{100\%} \\
    \midrule
    \textbf{Random Init.} & 6.59  & 19.81 & 19.14 & 34.76 & 16.96 & 24.96 \\
    \textbf{SlotCon} & 17.53 & 38.37 & 26.54 & 48.45 & 20.72 & 28.78 \\
     \midrule
    \textbf{DeCon-SL ($\alpha=0.25$)} & 17.06 & 39.94 & 26.81 & \textbf{48.85} & 20.56 & 29.39 \\
    \textbf{DeCon-ML-S ($\alpha=0$, drop=0.5)} & 17.82 & 39.40 & 26.82 & 48.53 & \textbf{21.00} & 29.16 \\
    \textbf{DeCon-ML-L  ($\alpha=0$, drop=0.5)} & \textbf{17.98} & \textbf{40.03} & \textbf{27.16} & 48.57 & 20.66 & \textbf{29.77} \\
    \bottomrule
    \end{tabular}%
    }
  \label{tab:agri}%
\end{table}%

\subsection{Ablation studies}

\Cref{tab:ablation} presents an ablation study of the components in our DeCon-ML-L framework. The results show that while pre-training with a decoder and deep-supervision provides a modest benefit to downstream fine-tuning performance on COCO, the main performance improvement arises from the use of channel dropout in conjunction with decoder deep supervision.  This combination yields the most substantial gains, highlighting the critical role of channel dropout when paired with deep supervision in our approach. 

\begin{table}[htbp]
  \centering
  \caption{Ablation of the different components of DeCon-ML-L and their impact on the downstream performance. $\Delta$ is the performance gap between the previous framework and the framework updated with the corresponding new component.}
  \scalebox{0.65}{
       \begin{tabular}{p{17.165em}cccc}
    \toprule
    \multicolumn{1}{c}{\multirow{2}[4]{*}{\textbf{Framework Improvement}}} & \multicolumn{2}{c}{\textbf{COCO Obj Det.}} & \multicolumn{2}{c}{\textbf{COCO Inst. Seg.}} \\
\cmidrule{2-5}    \multicolumn{1}{c}{} & \textbf{AP} & \boldmath{}\textbf{$\Delta$}\unboldmath{} & \textbf{AP} & \boldmath{}\textbf{$\Delta$}\unboldmath{} \\
    \midrule
    \multicolumn{1}{l}{SlotCon} & 40.81 & \multicolumn{1}{c}{\textbf{-}} & 36.80 & \multicolumn{1}{c}{\textbf{-}} \\
    \multicolumn{1}{l}{+FPN Decoder (Multi Level Loss)} & 40.85 & +0.04 & 36.83 & +0.03 \\
    \multicolumn{1}{l}{+Dropout = 0.5} & 41.10 & +0.25 & 37.05 & +0.22 \\
    +Enc-Dec Loss Weight ($\alpha$) Tuning (DeCon-ML-L) & 41.18 & +0.08 & 37.12 & +0.07 \\
    \bottomrule
    \end{tabular}%
}
  \label{tab:ablation}%
\end{table}%

\Cref{tab:alphadropout} shows DeCon-ML performance with different values of encoder loss weight $\alpha$, channel dropout, and number of decoder levels. A channel dropout probability of 0.5 provided the best downstream performance for DeCon-ML-L. Hence, this value was chosen to ablate $\alpha$. An $\alpha$ value of 0 gave the best downstream performance meaning that the decoder loss can replace the encoder one and pre-train efficiently an encoder on its own. We also ablated with number of decoder levels in DeCon-ML-L where 4 decoder levels yielded the best averaged performance across tasks. For DeCon-SL, an $\alpha$ value of 0.25 resulted in the best downstream performance, suggesting that having an encoder loss at the bottleneck is still beneficial to the pre-training in the absence of skip connections.

\begin{table}[htbp]
  \centering
  \caption{Ablation of channel dropout, encoder loss weight ($\alpha$), and decoder levels in DeCon, pre-trained on COCO for 800 epochs and evaluated by fine-tuning on downstream tasks.}
  \scalebox{0.43}{
    
    \begin{tabular}{ccccccc||ccccccc}
    \toprule
    \multicolumn{1}{c}{\multirow{3}[4]{*}{\textbf{\shortstack{Abla\\tion}}}} & \multicolumn{1}{c}{\multirow{3}[4]{*}{\textbf{\shortstack{Frame\\work}}}} & \multicolumn{1}{c}{\multirow{3}[4]{*}{\textbf{\shortstack{Deco\\der}}}} & \multicolumn{1}{c}{\multirow{3}[4]{*}{\boldmath{}\textbf{$\alpha$}\unboldmath{}}} & \multicolumn{1}{c}{\multirow{3}[4]{*}{\textbf{Drop.}}} & \textbf{Obj. Det.} & \textbf{Inst. Seg.} & 
    \multicolumn{1}{c}{\multirow{3}[4]{*}{\textbf{\shortstack{Abla\\tion}}}} &
    \multicolumn{1}{c}{\multirow{3}[4]{*}{\textbf{\shortstack{Frame\\work}}}} & \multicolumn{1}{c}{\multirow{3}[4]{*}{\textbf{\shortstack{Deco\\der}}}} & \multicolumn{1}{c}{\multirow{3}[4]{*}{\boldmath{}\textbf{$\alpha$}\unboldmath{}}} & \multicolumn{1}{c}{\multirow{3}[4]{*}{\textbf{\shortstack{Dec\\Levels}}}}  & \multicolumn{2}{c}{\textbf{Sem. Seg.}} \\
          &       &       &       &       & \textbf{COCO} & \textbf{COCO} &       &       &  &  &   & \textbf{VOC} & \textbf{City} \\
\cmidrule{6-7}\cmidrule{13-14}          &       &       &       &       & \textbf{AP} & \textbf{AP} &  &      &   &    &       & \textbf{mIoU} & \textbf{mIoU} \\
    \midrule
    \multicolumn{1}{c}{\multirow{4}[2]{*}{\begin{sideways}\textbf{Dropout}\end{sideways}}} & \multicolumn{1}{c}{\multirow{8}[4]{*}{\begin{sideways}\textbf{\shortstack{DeCon-ML-L \\ (4 decoder levels)}}\end{sideways}}} & \multicolumn{1}{c}{\multirow{8}[4]{*}{\begin{sideways}\textbf{FPN}\end{sideways}}} & \multirow{4}[2]{*}{0.50} & 0     & 40.85 & 36.83
   & \multicolumn{1}{c}{\multirow{4}[2]{*}{\begin{sideways}\textbf{Dec. Level}\end{sideways}}} & \multicolumn{1}{c}{\multirow{4}[2]{*}{\shortstack{\textbf{DeCon-ML-L}\\(Drop = 0.5)}}}     & \multicolumn{1}{c}{\multirow{4}[2]{*}{\begin{sideways}\textbf{FPN}\end{sideways}}}    &  \multirow{3}[2]{*}{0}     & 2     & 72.70 & 76.07  \\
          &       &       &       & 0.25  & 40.95 & 36.97 & &      &      &      & 3     & \textbf{73.14} & 76.10 \\
          &       &       &       & 0.50  & \textbf{41.10} & \textbf{37.05} &      &&      &      & 4     & 72.92 & \textbf{76.45} \\
          &       &       &       & 0.75  & 41.02 & 37.05 &&      &     & -     & -     & - & - \\
\cmidrule{1-1}\cmidrule{4-14}    \multicolumn{1}{c}{\multirow{4}[2]{*}{\begin{sideways}\boldmath{}\textbf{$\alpha$ }\unboldmath{}\end{sideways}}} &       &       & 0.75  & \multirow{4}[2]{*}{0.50} & 40.98 & 36.90 & \multicolumn{1}{c}{\multirow{4}[2]{*}{\begin{sideways}\textbf{$\alpha$}\end{sideways}}} & \multicolumn{1}{c}{\multirow{4}[2]{*}{\textbf{\shortstack{DeCon-SL \\(No Drop)}}}} & \multicolumn{1}{c}{\multirow{4}[2]{*}{\begin{sideways}\textbf{FCN}\end{sideways}}} & 0.75& -  & 71.98 & 76.18 \\
          &       &       & 0.50  &       & 41.10 & 37.05 &&       &       & 0.50 & - & 72.42 & 75.62 \\
          &       &       & 0.25  &       & 41.05 & 37.13 &&       &       & 0.25 & - & \textbf{73.01} & \textbf{76.21} \\
          &       &       & 0     &       & \textbf{41.18} & \textbf{37.11} &&       &       & 0  & -   & 72.17 & 75.71 \\
    \bottomrule
    \end{tabular}%
    }
  \label{tab:alphadropout}%
\end{table}%

%% file: our_sections/5_discussion_wacv.tex
\section{Discussion}
\label{sec:discussion}

In this paper, we introduced a contrastive framework adaptation by incorporating a decoder contrastive loss to an existing framework. We showed that our proposed SSL adaptation achieved SOTA performance in object detection, instance and semantic segmentation tasks when pre-trained on COCO+ and ImageNet-1K datasets, and in most tasks when pre-trained on COCO (\cref{tab:main}). This adaptation enhances the representation of the pre-trained encoder, improving its transfer performance for dense downstream tasks, even in tasks beyond object detection, instance and semantic segmentation, regardless of the homogeneity between pre-training and fine-tuning decoders.


%
%

%
%
\begin{figure}[h]
    \centering
    \includegraphics[width=0.47\textwidth]{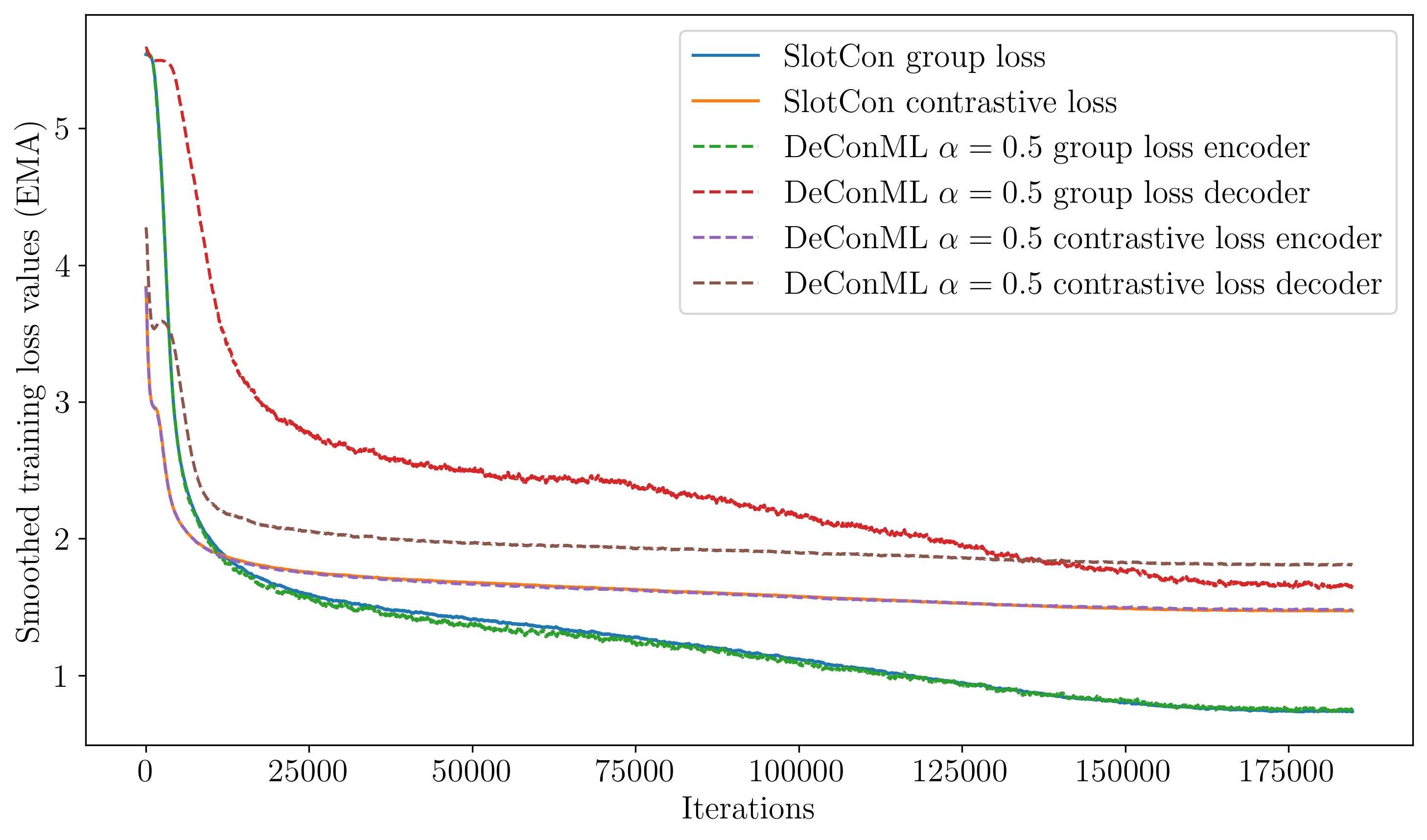}
    \caption{DeCon-ML-L and SlotCon pre-training loss dynamics.}
    \label{fig:losses}
\end{figure}
\begin{figure}[h]
    \centering
    \includegraphics[width=0.4\textwidth]{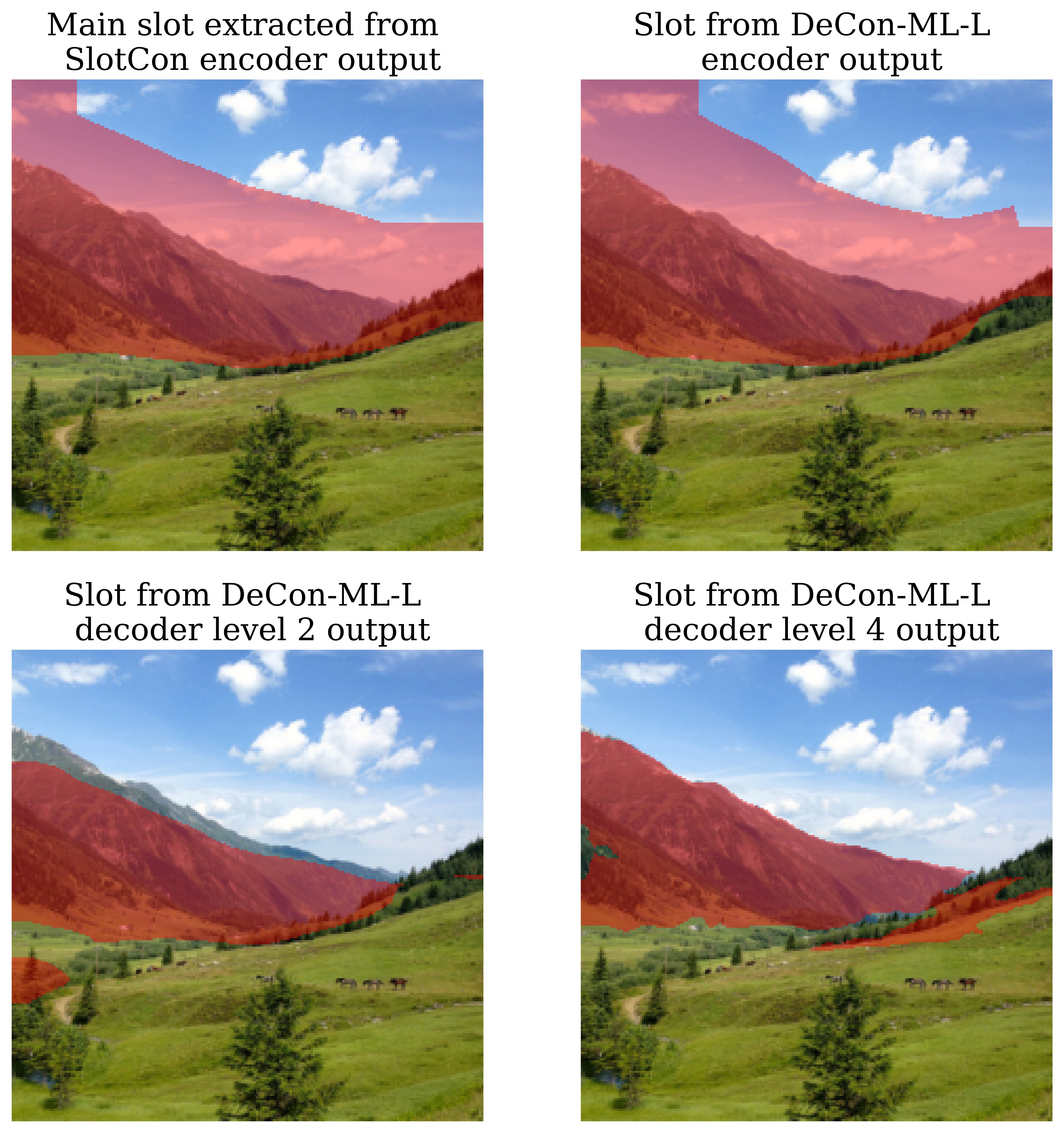}
    \caption{Slots, as defined in SlotCon~\cite{slotcon_paper_ssl}, learned at different outputs of our architecture. The top left image from COCO validation dataset is overlayed with the slot that was the most represented in original SlotCon's encoder output feature map, resized to the image's shape. Other slots displayed are the slots that had the biggest overlap with SlotCon's main encoder output slot. More details on the slots creation are available in Supplementary Material 6.}
    \label{fig:slots_different_features}
\end{figure}

%
We introduced a combined loss where encoder and decoder losses contribute to the total loss in a weighted manner. 
\Cref{fig:losses} illustrates losses behavior across different pre-trainings when both the encoder and decoder are used. As shown, decoder pre-training does not affect the encoder loss dynamics, supporting the fact that the encoder and decoder loss terms are conceptually non-competing. We also observed that increasing the weight for the decoder loss compared to that of the encoder enhances the performance of the pre-trained models. This could be attributed to the non-competing nature of the encoder and decoder losses compared to approaches that use a reconstruction loss for a decoder and a different loss for the encoder in an encoder-decoder architecture~\cite{competing_loss_contrast_gen}. Additionally, \cref{fig:slots_different_features}  presents the slots learned by SlotCon encoder and by the different decoder levels of DeCon-ML. It demonstrates that the decoder learns features that capture similar concepts as the encoder’s, while being more spatially precise. 

Moreover, we noticed that the encoder loss is optional when pre-training with DeCon-ML, demonstrating the efficiency of the decoder deep supervision and channel dropout combination to ensure a comprehensive pre-training of both encoder and decoder. The lack of skip-connections in DeCon-SL with FCN decoder makes the combined encoder-decoder loss preferable to ensure the best pre-training. In Supplementary Material Section 7, we provide experiments on using DeCon-SL with an FPN decoder. We observed that with an FPN decoder, dropout does not provide the same benefits in DeCon-SL compared to DeCon-ML. Hence, channel dropout should only be applied to the different encoder levels when pre-training with a skip-connected decoder and decoder deep supervision (DeCon-ML). Framework-specific hyperparameter tuning of the decoder auxiliary layers could also increase the fine-tuning performance (Supplementary Material Section 8).\par

%

We demonstrated that our method is effective with different encoders. Specifically, pre-training a ConvNeXt-Small backbone, which is both more efficient and larger than the ResNet-50 backbone, also resulted in superior downstream performance when integrated with a decoder in the DeCon-ML-L framework. These results attest to the scalability and practical utility of the proposed approach for pre-training modern backbone architectures. Additionally, \cref{tab:generative} shows that a ConvNeXt-S backbone pre-trained with DeCon-SL for dense tasks outperforms larger ViT-based SSL methods, despite using fewer pre-training epochs.

DeCon was also used to successfully adapt three different SSL frameworks using different concepts (pixel level VS object/group level) to learn local information (\cref{tab:densecl}). This suggests that the proposed DeCon adaptation could be used in other SSL contrastive methods targeting dense tasks.\par

\begin{table}[htbp]
  \centering
  \caption{Pre-training parameter count and computational cost for the different architectures. We study various pre-training setups on the COCO dataset using the SlotCon framework. Module sizes are given in millions of parameters. GPU-days are calculated on H100 80GB GPUs.}
  \scalebox{0.55}{ 
    \begin{tabular}{ccccccc} 
    \toprule
    \textbf{Pre-training} & \textbf{Enc. size} & \textbf{Dec. size} & \textbf{\shortstack{Enc. Aux.\\Layers Size}} & \textbf{\shortstack{Dec. Aux.\\Layers Size}} & \textbf{\shortstack{Full Archi.\\Size}} & \textbf{\shortstack{GPU-\\days}}  \\
    \midrule
    SlotCon & 23.51 x 2 & 0     & 71.51 & 0     & 118.52 & 1.53 \\
    DeCon-SL-FCN ($\alpha$ = 0.25) & 23.51 x 2 & 5.31 x 2 & 71.51 & 56.83 & 185.97 & 1.77  \\
    DeCon-ML-S ($\alpha$ = 0) & 23.51 x 2 & 3.35 x 2  & 0 & 59.08 & 110.03 & 1.78   \\
    DeCon-ML-L ($\alpha$ = 0) & 23.51 x 2 & 3.35 x 2  & 0 & 227.31 & 281.02 & 11.00   \\
    
    \bottomrule
    \end{tabular}%
    }
  \label{tab:parameter}%
\end{table}%

DeCon adaptation of SSL frameworks yields clear advantages on small-scale datasets and out-of-domain dense prediction tasks, outperforming encoder-only baselines (\cref{tab:ood}, \cref{tab:agri}). These gains persist even when pre-training on the small target datasets, or when fine-tuning with only a fraction of labeled data. This highlights DeCon's value in domains with limited annotations, like medical and agricultural applications.

Performance of deep learning models can vary across multiple runs due to inherent stochasticity. To mitigate the impact of randomness while considering computational resource constraints, each experiment was conducted three times, with all fine-tunings for a specific downstream task performed on similar hardware. The performance improvements achieved with our DeCon framework compared to the original framework are statistically significant, as evidenced by large Cohen’s $d$ values ($>0.8$) (in Supplementary Material Tab. S6), exceeding two standard deviations (see Supplementary Material Section 9), and $p$-values less than 0.05 (0.012 for COCO object detection and 0.048 for instance segmentation) from Wilcoxon signed-rank tests. Details on the computation of these values are provided in Supplementary Material Section 10. Furthermore, the performance gains, shown in Supplementary Material Tab. S6, are comparable to those reported in prior studies.

A limitation of our approach is its high memory and computational demand. However, as shown in~\cref{tab:main}, when GPU access is limited, reducing the number of decoder levels and tuning the decoder projector layers can decrease the parameter count and training cost of DeCon-ML (e.g., DeCon-ML-S in~\cref{tab:parameter}) while still enhancing pre-training performance. It is also notable that DeCon-SL and DeCon-ML-S have comparable training costs as their baseline.

Brempong et al. \cite{Brempong_2022_CVPR} followed a staged approach, first pre-training the encoder with supervision on annotated datasets, then pre-training the decoder with denoising. However, this multi-stage process relies on labeled data. It could also result in suboptimal feature alignment and increased training complexity. 
Future works could involve extending our proposed DeCon adaptation to offer a multi-stage continual pre-training in a contrastive manner. We also plan to adapt DeCon to ViT architectures in the future.

%% file: our_sections/6_conclusion.tex
\section{Conclusion}
\label{sec:conclusion}
In this paper, we introduced DeCon---a novel contrastive framework adaptation that enhances the performance of SSL approaches for various dense predictions tasks. We showed that this adaptation improves the encoder's representation significantly. 
We proposed two variants of DeCon: DeCon-SL and DeCon-ML. The former introduces a decoder contrastive loss, and the latter extends DeCon-SL by introducing channel dropout and decoder deep supervision to maximize the encoder's pre-training power. By integrating the proposed contrastive joint encoder-decoder pre-training strategy into an existing framework, we show that DeCon consistently outperforms or matches state-of-the-art methods across diverse dense prediction tasks, highlighting the effectiveness of this unified pre-training approach.

%% file: our_sections/acknowledgments.tex
\section*{Acknowledgments}

We extend our sincere gratitude to the University of Calgary VPR Office for their generous support through VPR Catalyst Grants, which was instrumental in making this research possible. We also acknowledge the support of the Natural Sciences and Engineering Research Council of Canada (NSERC Discovery Grant RGPIN-2024-04966) and Alberta Innovates. This work was partly supported by Alberta Innovates Graduate Student Scholarship and T. Chen Fong Doctoral Research Excellence Scholarship in Imaging Science.
Computing resources were provided by the University of British Columbia Cloud Innovation Centre, powered by Amazon Web Services \url{https://cic.ubc.ca/}. This research was also partly enabled by support provided by the Digital Research Alliance of Canada \url{https://alliancecan.ca/}, and the Research Computing Services group at the University of Calgary \url{https://rcs.ucalgary.ca/RCS_Home_Page}.

%% file: supplementary.tex
\section{Decoder architecures}

\noindent\textbf{Fully Convolutional Network:}
To match previous evaluation with mmseg framework, we implement a fully convolutional Network (FCN) as a two 3 × 3 convolutional block of 256 channels (dilation set to 6) with batch normalization and ReLU activations.

\noindent\textbf{Feature Pyramid Network:}
Feature Pyramid Network (FPN) C5 implementation is customized from the Detectron2 library~\cite{detectron2_package}. The architecture has four lateral convolutional layers. Each layer uses a $1\times1$ convolution to reduce the ResNet-50 outputs from 2048, 1024, 512, and 256 channels down to 256 channels. We then sum these lateral outputs in a bottom-up manner and pass the combined output through a $3\times3$ convolution at each decoder level to produce a 256-channel decoder output. 

For pre-training with DeCon-ML under deep supervision, we apply a $3 \times 3$ convolution at each of the four decoder levels. The resulting outputs are then passed to separate auxiliary layers, yielding four distinct decoder losses. Note that, deep supervision loss is only calculated in FPN scenario where lateral connections are present.

\section{Implementation details }

\noindent\textbf{DenseCL pre-training:}
For DenseCL, the two encoders, along with their respective auxiliary layers---i.e., the global and dense projection heads---are kept the same as well as the independent dictionaries for global and dense losses. Additionally, we introduce two decoders, which take their respective encoders' outputs as inputs. We also replicate the encoder's dense and global projection heads and dictionaries separately for the two decoders. As for SlotCon, the input dimension of the decoder projectors is 256 instead of 2048 for the encoder. The size of the encoders and decoders dictionaries size is set to 16384. Other hyper parameters follow the official implementation. We follow DenseCL by updating the first branch of the architecture through Backpropagation, while the second branch is updated using an EMA of the first one.

\noindent\textbf{PixPro pre-training:}
For PixPro, the two encoders, along with their respective auxiliary layers---i.e., the projectors and pixel to propagation modules---are kept the same. Additionally, we introduce two decoders, which take their respective encoders' outputs as inputs. We also replicate the encoders' projection heads and propagation module for the decoders. As for SlotCon, the input dimension of the decoder projectors is 256 instead of 2048 for the encoder. We follow PixPro by updating the first branch of the architecture through backpropagation, while the second branch is updated using an EMA of the first one. Hyper parameters follow the official implementation.



\noindent\textbf{REFUGE and ISIC pre-training:}
We pre-train SlotCon and a DeCon-SL adaptation of SlotCon with an FCN decoder on two small-scale datasets: ISIC 2017 \cite{dataset_isic} and REFUGE \cite{dataset_refuge}. More details on these datasets are provided in the "Semantic segmentation on out-of-domain datasets" Section. For ISIC dataset, we pre-train for 800 epochs with a batch size of 256 and a base learning rate of 1.0 linearly scaled with the batch size. For REFUGE, we pre-train for 2400 epochs with a batch size of 192 and the same learning rate as for ISIC. These pre-trainings are performed with one NVIDIA A6000 GPU with 48GB of memory.

\noindent\textbf{Object detection and instance segmentation:} We performed object detection and instance segmentation on COCO 2017 dataset~\cite{coco_dataset}. We fine-tune a Mask R-CNN with FPN backbone for 90000 iterations on the COCO \textit{train2017} dataset  and evaluate on COCO \textit{val2017} split following SlotCon~\cite{slotcon_paper_ssl}. We initialize the network with a pre-trained ResNet-50 or a ConvNeXt-Small encoder along with a randomly initialized decoder, and fine-tune end-to-end in all cases with a batch size of 16. The base learning rate is set to 0.02 for ResNet-50 and to 0.0002 for the ConvNeXt-Small encoder. The ROI mask head uses four convolution layers and the ROI box head uses two fully connected layers. All the aforementioned experiments are performed using one NVIDIA A100 GPU with 80GB memory.

\noindent\textbf{In-domain semantic segmentation:}
We performed semantic segmentation on Pascal VOC~\cite{voc_dataset},  Cityscapes~\cite{cityscape_dataset}, and ADE20K~\cite{ade} datasets following SlotCon~\cite{slotcon_paper_ssl} and PixCon~\cite{pixcon_paper_ssl}. A ResNet-50 encoder with a two-layer FCN decoder, similar to the FCN decoder used during pre-training, is used. We also fine-tuned a ConvNeXt-Small backbone with a two-layer FCN decoder on Pascal VOC dataset. For ADE20K, two decoder heads are used: a one-layer FCN auxiliary decoder along with a two-layer FCN decoder following SlotCon \cite{slotcon_paper_ssl}. For Pascal VOC, we train on VOC \textit{train\_aug2012} set for 30000 iterations and evaluate on VOC \textit{val2012} set. For Cityscapes, we fine tune on the \textit{train\_fine} set for 90000 iterations and evaluate on the \textit{val\_fine} set. For ADE20k with ResNet-50-FCN, we fine tune on the \textit{training} set for 80000 iterations and evaluate on the \textit{validation} set. In all the cases, batch size was 16. In Pascal VOC tasks with ResNet-50-FCN, the models were optimized using SGD with a learning rate of 0.003, momentum of 0.9, and weight decay of 0.0001 with a step learning rate policy to reduce the learning rate by a factor of 0.1 at 21,000 and 27,000 iterations. For Cityscapes semantic segmentation with ResNet-50-FCN, the models were optimized with an SGD optimizer, a learning rate of 0.01, momentum of 0.9, and weight decay of 0.0001, while the learning rate is reduced by a factor of 0.1 at 63,000 and 81,000 iterations following a step policy. In ADE20K task with ResNet-50 FCN, the models were optimized with an SGD optimizer, a learning rate of 0.01, momentum of 0.9, and weight decay of 0.0005, with a polynomial learning rate decay that reduced the learning rate to 1e-4 over iterations.
All the aforementioned experiments are performed using one NVIDIA A100 GPU with 80GB (Cityscapes) or 40GB (Pascal VOC and ADE20K) memory.   

To compare performance with ViT-based methods, we have fine-tuned a ConvNeXt-Small backbone (pre-trained on ImageNet-1K for 250 epochs using DeCon-SL framework with FCN decoder) using an UPerNet \cite{upernet} decoder on ADE20K for 160000 iterations. An FCN auxiliary decoder was also used along with it. The model was optimized with AdamW optimizer, a learning rate of 0.0001, betas set to (0.9, 0.999), and a weight decay of 0.05. The parameter-wise learning rate decay follows a stage-wise scheme with a decay rate of 0.9 over 12 layers. The learning rate schedule consists of two phases: a linear warm-up from a factor of 1e-6 for the first 1500 steps, followed by polynomial decay from step 1500 to 160000, with a minimum learning rate of 0.0.


\noindent\textbf{Semantic segmentation on out-of-domain datasets:}
To evaluate the pre-trained model's generalizability to out-of-domain semantic segmentation tasks, we fine-tuned a ResNet-50 encoder coupled with an FCN decoder on REFUGE~\cite{dataset_refuge} and ISIC 2017~\cite{dataset_isic} dataset. We also fine-tuned a ResNet-50 encoder with a deeplabv3+ decoder~\cite{deeplabv3plus} and an FCN auxiliary decoder head on PlantSeg~\cite{plantseg}. We evaluate generalizability to out-of-domain datasets across different training set sizes, using 5\%, 25\%, and 100\% of randomly selected samples for the REFUGE and ISIC 2017 tasks. For the PlantSeg task, we further assess generalization by fine-tuning with 10\% of randomly selected training samples.

REFUGE dataset contains 1200 retinal images in total representing 3 different classes: optic disk, optic cup, and background, divided into 400 training, 400 validation, and 400 testing samples. We train a ResNet-50-FCN network end-to-end for 80000 iterations with a batch size of 16 on the \textit{training} setusing one NVIDIA A100 GPU (40GB). The learning rate starts at 0.01 and decreases with a polynomial scheduler with a power of 0.9. We evaluate the model on the \textit{test} set using the last iteration checkpoint and report the results.


ISIC 2017 is a skin-lesion segmentation dataset which consists of 2000 training, 150 validation, and 600 testing images. We fine-tune the same ResNet-50-FCN architecture on the \textit{training} set for 24000 iterations on one NVIDIA V100 with 32GB of memory. Hyper parameters are the same as for REFUGE. We evaluate on the \textit{validation} set every 500 iterations, select the best checkpoint and report the results of this checkpoint on the \textit{test} set. 

PlantSeg is a large-scale agricultural dataset for plant disease segmentation with 150 classes, comprising 11,458 images (9,163 for training and 2,295 for testing) across 34 plant varieties and 115 disease types. We fine-tune a ResNet-50-DeepLabv3+ \cite{deeplabv3plus} using SGD with a learning rate of 0.01, momentum 0.9, and weight decay 0.0005 on a single NVIDIA A100 GPU (40GB). The learning rate follows a polynomial decay schedule (power 0.9, minimum 1e-4) over 160,000 iterations. Evaluation is performed on the \textit{test} set using the final checkpoint.

\noindent\textbf{Object detection on out-of-domain datasets:}
To evaluate generalization in object detection, we fine-tuned our pre-trained encoders on the PlantDoc \cite{plantdoc} and Detecting Diseases \cite{detecting-diseases_dataset} datasets, using 10\% and 100\% of their training samples. The PlantDoc dataset~\cite{plantdoc} includes 2,569 images spanning 13 plant species and 30 object detection categories, covering both healthy and diseased plants. It contains 8,851 labeled instances, split into 2,328 training and 239 testing images. The Detecting Disease dataset consists of 5,493 leaf images across 13 disease categories, divided into 2,904 training, 1,416 validation, and 1,163 test images, with the test set used for evaluation.

For both of the tasks, we fine-tuned a Faster R-CNN framework with a ResNet-50 FPN backbone. These models are implemented using Detectron2~\cite{detectron2_package} and fine-tuned end-to-end. The total training schedule consists of 20000 iterations with 0.02 learning rate, learning rate decay steps at 12000 and 16000 iterations, and batch size of 4. We evaluated the model at every 1000 iterations, and reported the result with the best checkpoint.

\noindent\textbf{Panoptic segmentation, keypoint detection and dense pose estimation:} 
For panoptic segmentation and keypoint detection on COCO, we adopt the standard configurations provided in the Detectron2 package. Keypoint detection is performed using a Faster R-CNN configuration, while panoptic segmentation employs a Mask R-CNN configuration. Both tasks are trained on the COCO \textit{train2017} set with a learning rate of 0.02, a batch size of 16, and 90,000 iterations. For panoptic segmentation, the ROI box head is modified to align with the SlotCon object detection architecture, consisting of four convolutional layers followed by a fully connected layer. Model performance is evaluated on the COCO \textit{val2017} set using the last saved checkpoint.

Similarly we adapt the default detectron2 configuration for human dense pose estimation. We fine-tune a densepose R-CNN with a ResNet-50 encoder and an FPN decoder for 130000 iterations on COCO \textit{train2014}, with a batch size of 16 and a learning rate of 0.01. The ROI box Head is also adapted to match SlotCon object detection architecture with four convolution layers and one fully connected layer. We evaluate performance on the COCO \textit{val2014} using the last checkpoint saved.

\section{SlotCon adaptations}
\begin{figure*}[t]
  \centering

    \includegraphics[width=0.8\linewidth]{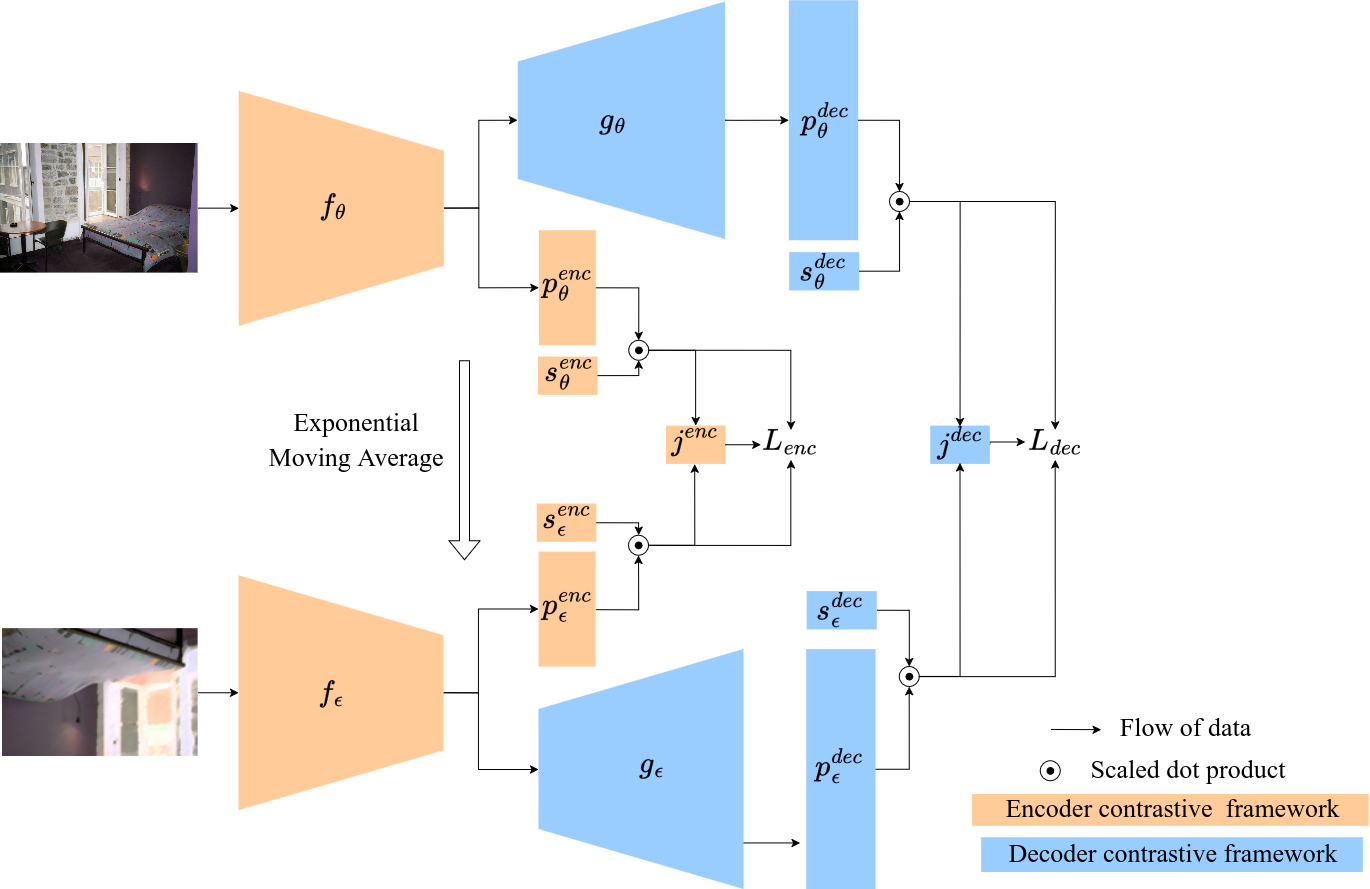}
   \caption{This figure illustrates the proposed DeCon-SL adaptation of SlotCon framework with an FCN decoder. $f_\theta$ and $f_\epsilon$ represent student and teacher encoders, respectively. $p^{enc}_{\theta} $ and $p^{enc}_{\epsilon}$ are the projector layers of the SSL frameworks of the student and teacher encoders, respectively.  $s^{enc}_{\theta} $ and $s^{enc}_{\epsilon} $ are the semantic grouping layers of the SSL frameworks of the student and teacher encoders, respectively. $j^{enc} $ is the encoder predictor slot. $g_\theta$ and $g_\epsilon$ represent student and teacher decoders, respectively. $p^{dec}_{\theta} $ and $p^{dec}_{\epsilon}$ are the projector layers of the SSL frameworks of the student and teacher decoders, respectively.  $s^{dec}_{\theta} $ and $s^{dec}_{\epsilon} $ are the semantic grouping layers of the SSL frameworks of the student and teacher decoders, respectively. $j^{dec} $ is the decoder predictor slot. }
   \label{fig:slotcon_fcn}
\end{figure*}

\begin{figure*}[t]
  \centering

    \includegraphics[width=1\linewidth]{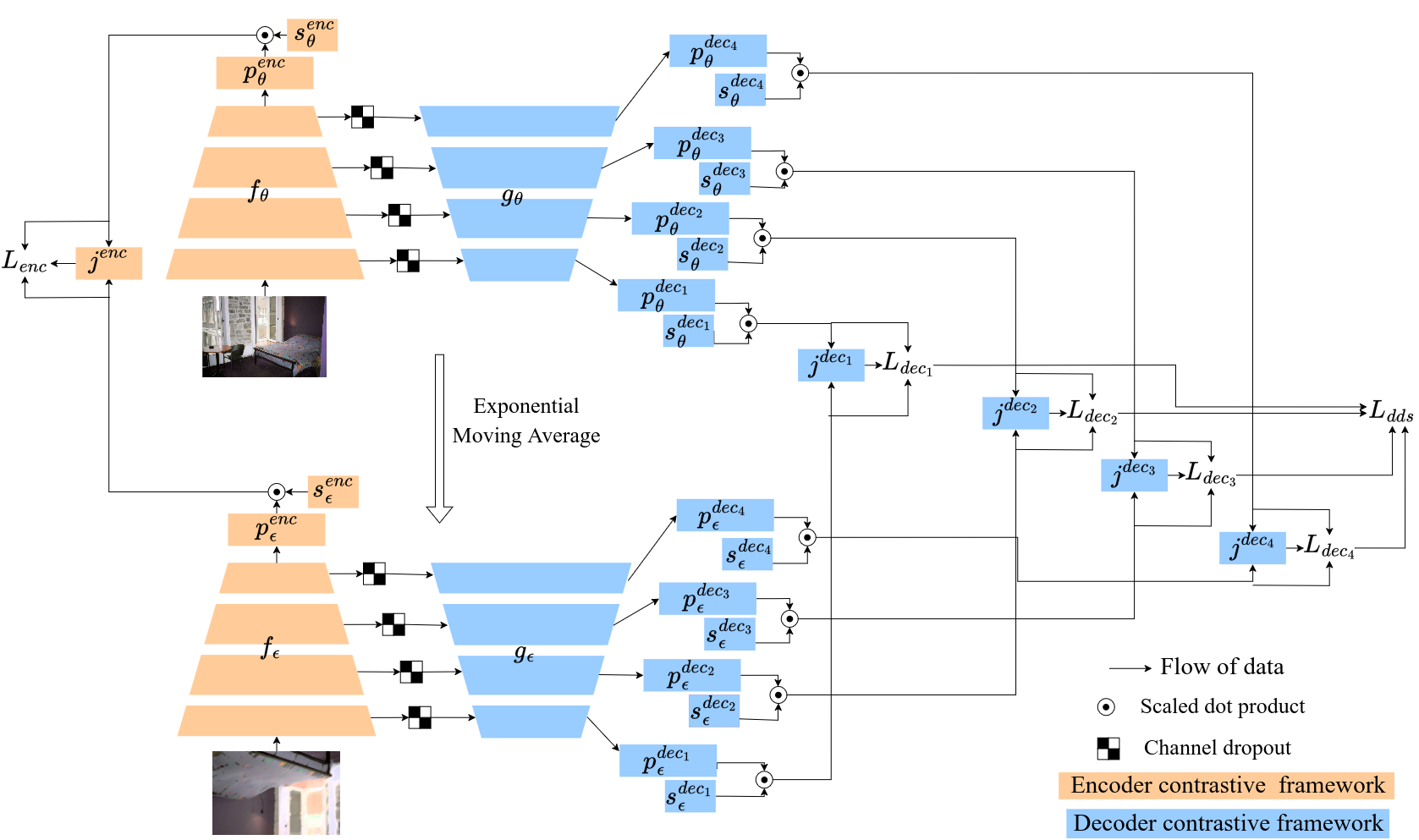}
   \caption{This schema illustrates the DeCon-ML-L adaptation of SlotCon framework with an FPN decoder. It also depicts the proposed decoder deep supervision and channel dropout. $f_\theta$ and $f_\epsilon$ represent student and teacher encoders, respectively. $p^{enc}_{\theta} $ and $p^{enc}_{\epsilon}$ are the projector layers of the SSL frameworks of student and teacher encoders, respectively.  $s^{enc}_{\theta} $ and $s^{enc}_{\epsilon} $ are the semantic grouping layers of the SSL frameworks of the student and teacher encoders, respectively. $j^{enc} $ is the encoder predictor slot. $g_\theta$ and $g_\epsilon$ represent student and teacher decoders, respectively. $p^{dec_i}_{\theta} $ and $p^{dec_i}_{\epsilon}$ are the projector layers of the SSL frameworks of the student and teacher decoders, respectively.  $s^{dec_i}_{\theta} $ and $s^{dec_i}_{\epsilon} $ are the semantic grouping layers of the SSL frameworks of the student and teacher decoders, respectively. $j^{dec} $ is the decoder predictor slot.}
   \label{fig:slotcon_fpn}
\end{figure*}

\begin{figure*}[t]
  \centering

    \includegraphics[width=1\linewidth]{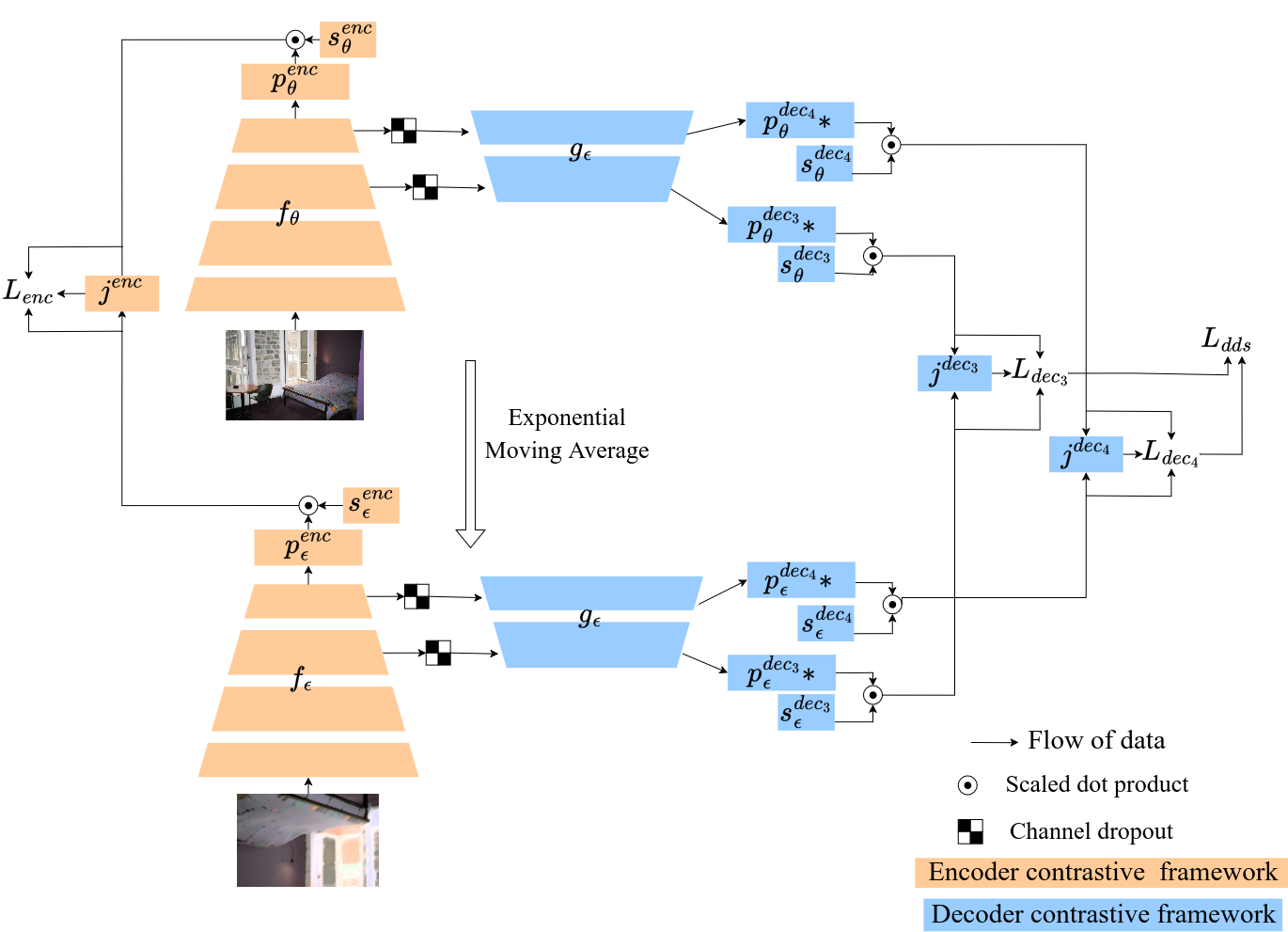}
   \caption{This schema illustrates the DeCon-ML-S adaptation of SlotCon framework with an FPN decoder. DeCon-ML-S is a smaller version of DeCon-ML-L where only the two first decoder levels are used and the decoder projector hidden dimension is reduced. * means the hidden dimension was altered from 4096 to 2048.}
   \label{fig:slotcon_fpn_small}
\end{figure*}

\Cref{fig:slotcon_fcn} presents the DeCon-SL adaption of SlotCon SSL framework to an encoder-decoder framework.
\Cref{fig:slotcon_fpn} and \Cref{fig:slotcon_fpn_small} illustrates how we used channel dropout and decoder deep-supervision for the DeCon-ML-L and DeCon-ML-S adaptations of SlotCon framework with an FPN decoder pre-training.

\section{Performance comparison with state of the art methods}
\Cref{tab:perf_comp} presents a performance comparison between our best models, DeCon-ML-L ($\alpha=0$, dropout$=0.5$) and DeCon-SL ($\alpha=0.25$) adaptations of SlotCon, and existing state-of-the-art methods. This table extends Tab. 1 from the main manuscript to better reflect the range of performances of the different SSL methods.

\begin{table*}[htbp]
  \centering
  \caption{Performance comparison with state-of-the-art SSL frameworks pre-trained on COCO, COCO+ and ImageNet-1K. DeCon-ML-L and DeCon-ML-S were pre-trained with FPN decoder  ($\alpha=0, dropout=0.5$) and DeCon-SL was pre-trained with FCN decoder ($\alpha=0.25$) adaptation of SlotCon. We only transferred the encoder for downstream tasks. We averaged fine-tuning results over three runs. $\dag$: Collected from PixCon paper. $\lozenge$ Collected from SlotCon Paper $\ddag$: Full re-implementation.}
  \scalebox{0.8}{
    \begin{tabular}{clcccccccccc}
    \toprule
    \multicolumn{1}{c}{\multirow{3}[4]{*}{\textbf{\shortstack{Pret.\\ Dataset}}}} & \multicolumn{1}{c}{\multirow{3}[4]{*}{\textbf{Framework}}} & \multicolumn{1}{c}{\multirow{3}[4]{*}{\textbf{Pret. Dec.}}} & \multicolumn{3}{c}{\textbf{Object Detection}} & \multicolumn{3}{c}{\textbf{Instance Segmentation}} & \multicolumn{3}{p{10.5em}}{\textbf{Semantic Segmentation}} \\
          &       &       & \multicolumn{3}{c}{\textbf{COCO}} & \multicolumn{3}{c}{\textbf{COCO}} & \textbf{VOC} & \textbf{City} & \textbf{ADE} \\
\cmidrule{4-12}          &       &       & \textbf{AP} & \textbf{AP50} & \textbf{AP75} & \textbf{AP} & \textbf{AP50} & \textbf{AP75} & \textbf{mIoU} & \textbf{mIoU} & \textbf{mIoU} \\
    \midrule
     & \textbf{Random init.$\lozenge$} & \textbf{-} & 32.8  & 50.9  & 35.3  & 29.9  & 47.9  & 32.0  & 39.5  & 65.3 & 29.4 \\
    \midrule
    \multirow{14}[2]{*}{\begin{sideways}\textbf{COCO}\end{sideways}}   & \textbf{MoCo-v2 (2020) $\lozenge$~\cite{mocov2_paper_ssl}} & \textbf{-} & 38.5  & 58.1  & 42.1  & 34.8  & 55.3  & 37.3  & 69.2  & 73.8 & 36.2 \\
          & \textbf{BYOL (2020) \dag~\cite{byol_paper_ssl}} & \textbf{-} & 39.5  & 59.4  & 43.3  & 35.6  & 56.6  & 38.2  & 70.2  & 75.3 & \textbf{-} \\
          & \textbf{MoCo-v2 + (2022) \dag~\cite{mocov2+_paper_ssl}} & \textbf{-} & 39.8  & 59.7  & 43.6  & 35.9  & 57.0  & 38.5  & 71.1  & 75.6 & \textbf{-}  \\
          & \textbf{ORL (2021) $\lozenge$~\cite{orl_paper}} & \textbf{-} & 40.3  & 60.2  & 44.4  & 36.3  & 57.3  & 38.9  & 70.9  & 75.6 & 36.7 \\
          & \textbf{PixPro (2021) $\lozenge$~\cite{pixpro_paper_ssl}} & \textbf{-} & 40.5  & 60.5  & 44.0  & 36.6  & 57.8  & 39.0  & 72.0  & 75.2 & 38.3 \\
          & \textbf{DetCon (2021) $\lozenge$~\cite{detcon_paper_ssl}} & \textbf{-} & 39.8  & 59.5  & 43.5  & 35.9  & 56.4  & 38.7  & 70.2  & 76.1 & 38.1 \\
          & \textbf{UniVIP (2022) \dag~\cite{univip_paper_ssl}} & \textbf{-} & 40.8  & \textbf{-} & \textbf{-} & 36.8  & \textbf{-} & \textbf{-} & - & - & - \\
          & \textbf{Odin (2022) \dag~\cite{odin_paper_ssl}} & \textbf{-} & 40.4  & 60.4  & 44.6  & 36.6  & 57.5  & 39.3  & 70.8  & 75.7 & \textbf{-} \\
          & \textbf{DenseCL (2021) $\lozenge$~\cite{densecl_paper_ssl}} & \textbf{-} & 39.6  & 59.3  & 43.3  & 35.7  & 56.5  & 38.4  & 71.6  & 75.8 & 37.1 \\
           & \textbf{DenseCL-D (2024) \cite{mindAug2024}} & \textbf{-} & 39.3  & 58.7  & 42.6  & 34.2  & 55.7    & 36.5  & -  & - & -\\
  & \textbf{SoCo-D (2024) \cite{mindAug2024} } & \textbf{-} & 40.3  & 60.1  & 44.0  & 35.1  & 56.9    & 37.6  & \textbf{-}  & \textbf{-} & \textbf{-}  \\
          & \textbf{PixCon-SR (2024) \dag ~\cite{pixcon_paper_ssl}} & \textbf{-} & 40.81 & 60.97  & 44.80  & 36.80 & 57.93  & 39.62 & \underline{72.95} & \textbf{76.62} & 38.0 \\
          & \textbf{Slotcon (2022) \ddag ~\cite{slotcon_paper_ssl}} & \textbf{-} & 40.81 & 60.95 & 44.37 & 36.80 & 57.98 & 39.54 & 71.50 & 75.95 & 38.57 \\
          & \textbf{DeCon-SL (SlotCon adapt.) [ours]} & \textbf{FCN} & 40.97 & 61.22 & 44.81 & 36.92 & 58.12 & 39.78 & \underline{73.01} & 76.21 & \textbf{38.81}\\
          & \textbf{DeCon-ML-S (SlotCon adapt.) [ours]} & \textbf{FPN} & 40.97 & 61.20 & 44.71 & 36.94 & 58.20 & 39.63 & 72.80 & 76.21 & 38.36 \\
          & \textbf{DeCon-ML-L (SlotCon adapt.) [ours]} & \textbf{FPN} & \textbf{41.18} & \textbf{61.38} & \textbf{44.91} & \textbf{37.12} & \textbf{58.35} & \textbf{39.94} & \underline{72.92} & 76.45 & 38.70 \\
    \midrule
    \multirow{7}[2]{*}{\begin{sideways}\textbf{COCO+}\end{sideways}} & \textbf{ORL (2021) \dag~\cite{orl_paper}} & \textbf{-} & 40.6  & \textbf{-} & \textbf{-} & 36.7  & \textbf{-} & \textbf{-} & \textbf{-} & \textbf{-} & \textbf{-} \\
          & \textbf{UniVIP (2022) \dag~\cite{univip_paper_ssl}} & \textbf{-} & 41.1  & \textbf{-} & \textbf{-} & 37.1  & \textbf{-} & \textbf{-} & \textbf{-} & \textbf{-} & \textbf{-} \\
          & \textbf{PixCon-SR (2024) \dag~\cite{pixcon_paper_ssl}} & \textbf{-} & 41.2  & \textbf{-} & \textbf{-} & 37.1  & \textbf{-} & \textbf{-} & 73.9  & \multicolumn{1}{c}{\underline{77.0}} & 38.8 \\
          & \textbf{Slotcon (2022) \ddag~\cite{slotcon_paper_ssl}} & \textbf{-} & 41.63 & 62.10 & 45.67 & 37.57 & 59.07 & 40.45 & 73.93 & 76.43 & 39.11 \\
          & \textbf{DeCon-SL (SlotCon adapt.) [ours]} & \textbf{FCN} & 41.86 & 62.43 & 45.73 & 37.75 & 59.40 & 40.48 & 74.46 & 76.65 & \textbf{39.25} \\
          & \textbf{DeCon-ML-L (SlotCon adapt.) [ours]} & \textbf{FPN} & \textbf{42.08} & \textbf{62.42} & \textbf{46.13} & \textbf{37.84} & \textbf{59.41} & \textbf{40.75} & \textbf{75.36} & \underline{77.00} & 39.04 \\
    \midrule
    \multirow{10}[2]{*}{\begin{sideways}\textbf{ImageNet-1K}\end{sideways}}  &  \textbf{Supervised $\lozenge$} & \textbf{-} & 39.7  & 59.5  & 43.3  & 35.9  & 56.6  & 38.6  & 74.4  & 74.6 & 37.9 \\
    \cmidrule{2-12}
    &  \textbf{MoCo-v2 (2020) $\lozenge$~\cite{mocov2_paper_ssl}} & \textbf{-} & 40.4  & 60.1  & 44.2  & 36.5  & 57.2  & 39.2  & 73.7  & 76.2 & 36.9 \\
          & \textbf{DetCo (2021) $\lozenge$~\cite{detco_paper_ssl} }& \textbf{-} & 40.1  & 61.0    & 43.9  & 36.4  & 58    & 38.9  & 72.6  & 76 & 37.8 \\
          & \textbf{InsLoc (2021) $\lozenge$~\cite{insloc_paper_ssl} }& \textbf{-} & 40.9  & 60.9  & 44.7  & 36.8  & 57.8  & 39.4  & 72.9  & 75.4 & 37.3\\
          & \textbf{DenseCL (2021) $\lozenge$~\cite{densecl_paper_ssl}} & \textbf{-} & 40.3  & 59.9  & 44.3  & 36.4  & 57    & 39.2  & 72.8  & 76.2 & 38.1 \\    
          & \textbf{PixPro (2021) $\lozenge$~\cite{pixpro_paper_ssl} }& \textbf{-} & 40.7  & 60.5  & 44.8  & 36.8  & 57.4  & 39.7  & 73.9  & \textbf{76.8} & 38.2 \\
          & \textbf{DetCon (2021) $\lozenge$~\cite{detcon_paper_ssl}} & \textbf{-} & 40.6  & \textbf{-} & \textbf{-} & 36.4  & \textbf{-} & \textbf{-} & 72.6  & 75.5 & \textbf{-} \\
          & \textbf{DINO (2021) \ddag~\cite{dino_paper_ssl}} & -     & 40.24 & 60.25  & 44.13 & 36.47 & 57.49 & 39.20 & 73.09 & 75.57 & 37.30  \\
          & \textbf{SoCo (2022) $\lozenge$~\cite{soco_paper_ssl}} & \textbf{-} & 41.6  & 61.9  & 45.6  & 37.4  & 58.8  & 40.2  & 71.9  & 76.5 & 37.8 \\
          & \textbf{Slotcon (2022) \ddag~\cite{slotcon_paper_ssl}} & \textbf{-} & 41.69 & 62.07 & 45.59 & 37.59 & 58.97 & 40.49 & 75.02 & 76.15 & 38.97 \\
          & \textbf{DeCon-ML-L (SlotCon adapt.) [ours]} & \textbf{FPN} & \textbf{41.80} & \textbf{62.12} & \textbf{45.73} & \textbf{37.73} & \textbf{59.08} & \textbf{40.68} & \textbf{75.40} & 76.51 &  \textbf{39.01}\\
    \bottomrule
    \end{tabular}%
    }
  \label{tab:perf_comp}%
\end{table*}%

\section{Encoder and Decoder transfer}
\cref{tab:encoder-decoder} presents the fine-tuning performance of the DeCon framework when transferring only the encoder versus transferring both the encoder and decoder. We observe that transferring both components does not consistently lead to a performance gain for the DeCon-SL and DeCon-ML-L adaptations of SlotCon. In contrast, a noticeable improvement is observed when both the encoder and decoder are transferred in the DeCon-SL adaptations of DenseCL and PixPro (see Tab. 4) in the main manuscript).
\begin{table}[htbp]
  \centering
  \caption{Performance of DeCon-SL ($\alpha=0.25$) and DeCon-ML-L ($\alpha=0$ and $dropout=0.5$) adaptations of SlotCon trained with a ResNet-50 backbone. The pre-trained encoder and decoder were used in the fine-tuning. All fine-tuning results were averaged over three runs.}
  \scalebox{0.55}{
    \begin{tabular}{ccclcccc}
    \toprule
    \multicolumn{1}{c}{\multirow{3}[4]{*}{\textbf{Pret. Dataset}}} & \multicolumn{1}{c}{\multirow{3}[4]{*}{\textbf{Framework}}} & \multicolumn{1}{c}{\multirow{3}[4]{*}{\textbf{Pret. Dec.}}} & \multicolumn{1}{c}{\multirow{3}[4]{*}{\textbf{Transfer}}} & \textbf{Obj. Det.} & \textbf{Inst. Seg.} & \multicolumn{2}{c}{\textbf{Sem. Seg.}} \\
          &       &       &       & \textbf{COCO} & \textbf{COCO} & \textbf{VOC} & \textbf{City} \\
\cmidrule{5-8}          &       &       &       & \textbf{AP} & \textbf{AP} & \textbf{mIoU} & \textbf{mIoU} \\
    \midrule
    \multirow{5}[6]{*}{COCO} & SlotCon & \multicolumn{1}{c}{-} & \multicolumn{1}{c}{Enc} & 40.81 & 36.80 & 71.50 & 75.95 \\
\cmidrule{2-8}          & \multicolumn{1}{c}{\multirow{2}[2]{*}{DeCon-SL}} & \multirow{2}[2]{*}{FCN} & \multicolumn{1}{c}{Enc} & \textbf{-} & \textbf{-} & \textbf{73.01} & 76.21 \\
          &       &       & Enc + Dec & \textbf{-} & \textbf{-} & 72.96 & \textbf{76.28} \\
\cmidrule{2-8}          & \multirow{2}[2]{*}{DeCon-ML-L} & \multirow{2}[2]{*}{FPN} & \multicolumn{1}{c}{Enc} & 41.18 & \textbf{37.12} & \textbf{-} & \textbf{-} \\
          &       &       & Enc + Dec & \textbf{41.21} & 37.11 & \textbf{-} & \textbf{-} \\
    \bottomrule
    \end{tabular}%
    }
  \label{tab:encoder-decoder}%
\end{table}%

\section{DeCon-ML-L slot selection}

Figure 3 in the main manuscript illustrates the visual representation of slots extracted from both encoder and decoder outputs. An image from the COCO \textit{val2017} dataset was used as input to SlotCon and DeCon-ML-L. Feature maps were extracted and projected using their respective projectors from the encoder bottleneck of both architectures, as well as from multiple decoder levels in the DeCon-ML-L framework. For each feature map (e.g., a 7×7 spatial map from the SlotCon encoder, a 14×14 map from level 2 of the DeCon-ML-L decoder or a 56×56 map from level 4 of the DeCon-ML-L decoder), each spatial location (pixel/vector) was assigned to the prototype with which it had the highest similarity, measured via dot product. The prototype most frequently assigned across all spatial positions was designated as the reference prototype (i.e., the slot) for that feature map.

To visualize this slot on the image, the dot product similarity between all feature vectors and the 256 learned prototypes was computed, resulting in a similarity map that was upsampled to the input image resolution (224×224). A voxel-wise argmax over the prototype dimension was used to determine prototype assignments, and a binary mask was generated by selecting only the pixels assigned to the reference prototype. This mask was overlaid on the image to highlight the most important slot/concept extracted from this image. A first mask was derived from the SlotCon encoder output and served as the reference slot.

For DeCon-ML-L, since separate prototypes—distinct from those of the SlotCon encoder—were learned at the encoder and at each decoder level, we repeated the same procedure to generate masks from corresponding feature map for all of the prototypes. For each of these, we identified the prototype whose mask showed the largest overlap with the SlotCon reference mask. This most overlapping prototype was considered to represent the same underlying concept, and its corresponding mask was overlaid on the reference image for comparison.

\section{DeCon-SL with an FPN decoder}

\Cref{tab:deconslfpn} presents the fine-tuning performance of the DeCon-SL adaptation of SlotCon with an FPN decoder. In this setting, two losses were used during the pre-training: the encoder loss and only the loss from the final layer of the FPN decoder. The results demonstrate that DeCon-SL consistently outperforms the original SlotCon when using an FPN decoder. As shown in the main manuscript, similar improvements were observed with an FCN decoder, suggesting that this approach is potentially generalizable to other decoder architectures. The greatest performance gains were observed when both the encoder and decoder were transferred. Notably, the DeCon-SL adaptation showed minimal sensitivity to the addition of dropout, even when the decoder included skip connections—a behavior that contrasts with the multi-level adaptation (DeCon-ML), where dropout had a more pronounced effect.

\begin{table*}[htbp]
  \centering
  \caption{ Performance of a DeCon-SL adaptation of SlotCon using a ResNet-50 encoder and an FPN decoder. The
architecture was pre-trained for 800 epochs on COCO, and then fine-tuned for object detection and instance segmentation tasks. The results are obtained with varying dropout rates (0 or 0.5) as well as transferring only the pre-trained encoder or both the pre-trained encoder and decoder. All
fine-tuning results were averaged over three runs.}
    \scalebox{0.8}{\begin{tabular}{ccclcccccccc}
    \toprule
    \multicolumn{1}{c}{\multirow{3}[4]{*}{\textbf{Pret. Dataset}}} & \multicolumn{1}{c}{\multirow{3}[4]{*}{\textbf{Framework}}} & \multicolumn{1}{c}{\multirow{3}[4]{*}{\textbf{Pret. Dec.}}} & \multicolumn{1}{c}{\multirow{3}[4]{*}{\textbf{Transfer}}} & \multicolumn{1}{c}{\multirow{3}[4]{*}{\boldmath{}\textbf{$\alpha$}\unboldmath{}}} & \multicolumn{1}{c}{\multirow{3}[4]{*}{\textbf{Dropout}}} & \multicolumn{3}{c}{\textbf{Object Detection}} & \multicolumn{3}{c}{\textbf{Instance Segmentation}} \\
          &       &       &       &       &       & \multicolumn{3}{c}{\textbf{COCO}} & \multicolumn{3}{c}{\textbf{COCO}} \\
\cmidrule{7-12}          &       &       &       &       &       & \textbf{AP} & \textbf{AP50} & \textbf{AP75} & \textbf{AP} & \textbf{AP50} & \textbf{AP75} \\
    \midrule
    \multirow{4}[6]{*}{COCO} & \multicolumn{1}{c}{SlotCon} & -     & \multicolumn{1}{c}{enc} & 1     & -     & 40.81 & 60.95 & 44.37 & 36.80 & 57.98 & 39.54 \\
\cmidrule{2-12}          & \multirow{3}[4]{*}{DeCon-SL} & \multirow{3}[4]{*}{FPN} & \multicolumn{1}{c}{enc} & 0.5   & 0     & 40.94 & 60.97 & 44.82 & 36.92 & 58.03 & 39.80 \\
          &       &       & \multicolumn{1}{c}{enc} & 0.5   & 0.5     & 40.90 & 61.23 & 44.64 & 36.91 & 58.30 & 39.67 \\
\cmidrule{4-12}          &       &       & enc+dec & 0.5   & 0     & 41.05 & 61.34 & 44.85 & 37.03 & 58.33 & 39.86 \\
    \bottomrule
    \end{tabular}%
    }
  \label{tab:deconslfpn}%
\end{table*}%

\begin{table*}[htbp]
  \centering
  \caption{Decoder specific hyperparameter tuning. We experiment on the number of prototypes to be used for the decoder pre-training in DeCon-SL adaptation. $\alpha$ is fixed to 0.5 and the number of prototypes used to compute the encoder loss is fixed to 256. We report the fine-tuning mIoU as an average over three runs. }
   \scalebox{1}{
 
    \begin{tabular}{ccccccclllll}
    \toprule
    \multirow{2}[4]{*}{\textbf{Dataset}} & \multirow{2}[4]{*}{\textbf{SSL}} & \multicolumn{2}{c}{\textbf{Loss}} & \multicolumn{2}{c}{\textbf{Transfer}} & \multicolumn{6}{c}{\textbf{Number of Dec. Prototypes}} \\
\cmidrule{3-12}          &       & \boldmath{}\textbf{$L_{enc}$}\unboldmath{} & \boldmath{}\textbf{$L_{dec}$}\unboldmath{} & \textbf{enc} & \textbf{dec} & \multicolumn{1}{l}{\textbf{0}} & \textbf{64} & \textbf{128} & \textbf{256} & \textbf{384} & \textbf{512} \\
    \midrule
    \multirow{3}[4]{*}{\textbf{VOC}} & \textbf{SlotCon} & \cmark & \xmark & \cmark & \xmark & \multicolumn{1}{l}{71.50} & \multicolumn{1}{c}{-} & \multicolumn{1}{c}{-} & \multicolumn{1}{c}{-} & \multicolumn{1}{c}{-} & \multicolumn{1}{c}{-} \\
\cmidrule{2-12}          & \multirow{2}[2]{*}{\textbf{DeCon-SL}} & \cmark & \cmark & \cmark & \xmark & -     & 72.10 & 71.95 & 72.42 & 72.43 & 72.32 \\
          &       & \cmark & \cmark & \cmark & \cmark & -     & 72.64 & 72.60 & 72.42 & \textbf{72.79} & 72.75 \\
    \midrule
    \multirow{3}[4]{*}{\textbf{City}} & \textbf{SlotCon} & \cmark & \xmark & \cmark & \xmark & \multicolumn{1}{l}{75.95} & \multicolumn{1}{c}{-} & \multicolumn{1}{c}{-} & \multicolumn{1}{c}{-} & \multicolumn{1}{c}{-} & \multicolumn{1}{c}{-} \\
\cmidrule{2-12}          & \multirow{2}[2]{*}{\textbf{DeCon-SL}} & \cmark & \cmark & \cmark & \xmark & -     & 75.67 & 75.79 & 75.67 & 75.87 & 75.97 \\
          &       & \cmark & \cmark & \cmark & \cmark & -     & 75.69 & 75.57 & 76.00 & \textbf{76.16} & 76.14 \\
    \bottomrule
    \end{tabular}%
    }
  \label{tab:hyperparameterSeg}%
\end{table*}%

\section{Decoder-specific hyperparameter tuning}

In all our previous experiments, hyperparameters used in the ``auxiliary layers'' of the encoder were replicated at the decoder level. \Cref{tab:hyperparameterSeg} shows that tuning these hyperparameters for the decoder could result in better downstream performance. Using 384 prototypes to compute the decoder pre-training loss results in better downstream performance for the DeCon-SL adaptation than using the encoder's parameter from the original SlotCon framework: 256 prototypes.

\section {Randomness of the result}
To account for variability arising from the stochasticity of training and the random initialization of the non-pre-trained components of the architecture, each downstream experiment was repeated three times. \Cref{tab:sig} reports the standard deviation of the fine-tuning performance. Furthermore, the Cohen’s d values in \cref{tab:coco+perf} of the main paper provide additional evidence that the observed performance improvements are statistically significant.

\begin{table*}[!htbp]
  \centering
  \caption{COCO and COCO+ pre-training downstream performance with standard deviation over 3 runs. Results for PixCon were obtained directly from the original PixCon publication, while SlotCon results were derived from full re-implementation. PixCon standard deviation of COCO+ pre-training was not provided due to unavailability of the checkpoint or reported standard deviation in the main paper.}
   \scalebox{0.8}{
    \begin{tabular}{lrrrr}
     
    \toprule
    \multicolumn{1}{c}{\multirow{2}[2]{*}{\textbf{Method}}} & \multicolumn{1}{l}{\textbf{COCO obj det.}} & \multicolumn{1}{l}{\textbf{COCO inst. Seg.}} & \multicolumn{1}{l}{\textbf{VOC sem. Seg.}} & \multicolumn{1}{l}{\textbf{City sem. Seg.}} \\
          & \multicolumn{1}{l}{\textbf{AP}} & \multicolumn{1}{l}{\textbf{AP}} & \multicolumn{1}{l}{\textbf{mIoU}} & \multicolumn{1}{l}{\textbf{mIoU}} \\
    \midrule
    \midrule
    \multicolumn{5}{c}{\textbf{(a) COCO Pre-training Performance}- ResNet50 backbone- Presented as mean over 3 runs ± std} \\
    \midrule
    SlotCon (2022) & \multicolumn{1}{l}{$40.81\pm0.16$} & \multicolumn{1}{l}{$36.80\pm0.18$} & \multicolumn{1}{l}{$71.50\pm0.27$} & \multicolumn{1}{l}{$75.95\pm0.23$} \\
    PixCon (2024) & \multicolumn{1}{l}{$40.81\pm0.09$} & \multicolumn{1}{l}{$36.84\pm0.11$} & \multicolumn{1}{l}{$72.95\pm0.29$} & \multicolumn{1}{l}{$76.62\pm0.10$} \\
    DeCon-ML (Comp. to SlotCon) & \multicolumn{1}{l}{$41.18\pm0.15$} & \multicolumn{1}{l}{$37.12\pm0.14$} & \multicolumn{1}{l}{$72.92\pm0.16$} & \multicolumn{1}{l}{$76.45\pm0.16$} \\
    \midrule
    \midrule
    \multicolumn{5}{c}{\textbf{(b) COCO+ Pre-training performance}- ResNet50 backbone - Presented as mean over 3 runs ± std} \\
    \midrule
    \midrule
    SlotCon (2022) & \multicolumn{1}{l}{$41.63\pm0.09$} & \multicolumn{1}{l}{$37.57\pm0.11$} & \multicolumn{1}{l}{$73.93\pm0.33$} & \multicolumn{1}{l}{$76.43\pm0.40$} \\
    PixCon (2024) & \multicolumn{1}{l}{41.2} & \multicolumn{1}{l}{37.1} & \multicolumn{1}{l}{73.9} & \multicolumn{1}{l}{77.0} \\
    DeCon-ML (Comp. to SlotCon) & \multicolumn{1}{l}{$42.08\pm0.08$} & \multicolumn{1}{l}{$37.84\pm0.05$} & \multicolumn{1}{l}{$75.36\pm0.21$} & \multicolumn{1}{l}{$77.00\pm0.53$} \\
    \bottomrule
    \end{tabular}%
    }
  \label{tab:sig}%
\end{table*}%

\section{Statistical Significance Tests}

\begin{table*}[!htbp]
  \centering
  \caption{COCO and COCO+ pre-training downstream performance improvement with DeCon-ML-L and Cohen's d value. PixCon performance was extracted from the original paper.}
   \scalebox{0.7}{

   \begin{tabular}{lcccc}
    \toprule
    \multicolumn{1}{c}{\multirow{2}[4]{*}{\textbf{Method}}} & \textbf{COCO obj det.} & \textbf{COCO Inst. Seg.} & \textbf{VOC Sem. Seg.} & \textbf{City Sem. Seg.} \\
\cmidrule{2-5}          & \textbf{$\Delta$AP} & \textbf{$\Delta$AP} & \textbf{$\Delta$mIoU} & \textbf{$\Delta$mIoU} \\
    \midrule
    \multicolumn{5}{c}{\textbf{ COCO Pre-training Performance Improvement $\Delta$ (calculated from the respective papers)}} \\
    \midrule
    \textbf{SlotCon (2022) (Comp. to prev. works)} & +0.50 & +0.40 & -0.40 & +0.10 \\
    \textbf{PixCon (2024) (Comp. to SlotCon)} & 0.00  & +0.06 & +1.30 & +0.51 \\
    \textbf{DeCon-ML-L (Comp. to SlotCon)} & +0.37 & +0.32 & +1.42 & +0.50 \\
    \midrule
    \textbf{Cohen’s d (DeCon-ML-L VS SlotCon) } & 2.35  & 1.94  & 6.37  & 2.51 \\
    \midrule
    \multicolumn{5}{c}{\textbf{COCO+ Pre-training Performance Improvement $\Delta$ (calculated from the respective papers)}} \\
    \midrule
    \textbf{SlotCon (2022) (Comp. to prev. works) } & +0.60 & +0.50 & Not avail. & Not avail. \\
    \textbf{PixCon (2024) (Comp. to SlotCon)} & -0.50 & -0.50 & -0.20 & +0.40 \\
    \textbf{DeCon-ML-L (Comp. to SlotCon) } & +0.45 & +0.27 & +1.43 & +0.57 \\
    \midrule
    \textbf{Cohen’s d (DeCon-ML-L VS SlotCon) } & 5.42  & 3.21  & 5.25  & 1.22 \\
    \bottomrule
    \end{tabular}%
    }
  \label{tab:coco+perf}%
\end{table*}%

In order to quantify the difference between mean metrics obtained when fine-tuning two different pre-trained models, we caclucated Cohen's $d$ value using the following formula:
\[
d = \frac{M_1 - M_2}{s_p}
\]

where \( M_1 \) and \( M_2 \) are the means of the two groups, and \( s_p \) is the pooled standard deviation given by:

\[
s_p = \sqrt{ \frac{(n_1 - 1)s_1^2 + (n_2 - 1)s_2^2}{n_1 + n_2 - 2} }
\]

Here, \( s_1 \) and \( s_2 \) are the standard deviations, and \( n_1 \), \( n_2 \) are the sample sizes of the two groups that we consider. In our case the sample size was 3 as we performed 3 runs for each fine-tuning experiments. The highly positive \( d \) values obtained when comparing SlotCon and DeCon-ML-L fine-tuning performance in~\cref{tab:coco+perf} indicates that our framework has a better average fine-tuning performance. \Cref{tab:coco+perf} also shows the performance improvements achieved with our DeCon framework compared to the original framework.


To evaluate the significance of the performance improvement of our framework over the original framework SlotCon, we performed a Wilcoxon signed-rank test on the Average Precision (AP) scores. We first computed per-image Average Precision (AP) scores for the three fine-tuning runs and stored the mean AP value of each image. We then perform the statistical test on these values. The p-values obtained, 0.012 for COCO object detection and 0.048 
for instance segmentation, show the significance of the performance improvement of the proposed framework.
